\RequirePackage{fix-cm}
\documentclass[10pt,twocolumn,letterpaper]{article}
\usepackage[pagenumbers]{cvpr}
\usepackage[english]{babel}
\usepackage{multirow}
\usepackage{microtype}
\usepackage[table]{xcolor}
\providecommand{\TODO}[1]{\textemdash}

\definecolor{cvprblue}{rgb}{0.21,0.49,0.74}
\usepackage[pagebackref,breaklinks,colorlinks,allcolors=cvprblue]{hyperref}
\usepackage{xcolor}
\usepackage{colortbl}
\usepackage{booktabs}
\usepackage{multirow}
\usepackage{graphicx}
\usepackage{makecell}
\definecolor{egcblue}{RGB}{55,105,150}
\definecolor{egcred}{RGB}{177,86,86}
\definecolor{overallgray}{gray}{0.95}

\newcommand{\egcbetter}[2]{%
  \ensuremath{\mathbf{#1}_{%
  \textcolor{egcblue}{\scriptsize\mathbf{(#2)}}}}}

\newcommand{\egcworse}[2]{%
  \ensuremath{\underline{#1}_{%
  \textcolor{egcred}{\scriptsize\mathbf{(#2)}}}}}

\def\confName{CVPR}
\def\confYear{2027}
\title{Allocentric Memory, Egocentric Acting: \\ A Dual-Frame VLA for Urban Aerial Navigation}
\author{
Haotian Xu$^{1,2}$ \quad
Yue Hu$^{1,2,*}$ \quad
Zhengqiu Zhu$^{1,2}$ \quad
Tuo Feng$^{3}$ \\
Wenguan Wang$^{3}$ \quad
Quanjun Yin$^{1,2}$ \\[0.6em]
$^{1}$National Key Laboratory of Digital Intelligent Modeling and Simulation\\
$^{2}$College of Systems Engineering, National University of Defense Technology\\
$^{3}$Zhejiang University\\[0.3em]
{\tt\small huyue11@nudt.edu.cn}\\
{\small $^{*}$Corresponding author}
}

\begin{document}
\maketitle

\begin{abstract}
Language-goal aerial navigation requires a UAV to reach a destination specified in natural language using observations collected during flight. A key challenge is linking changing egocentric perception with spatial evidence anchored to stable locations for long-range navigation. We propose \textbf{CityNav-VLA}, a dual-frame vision-language-action model for urban navigation integrating dynamic egocentric context with persistent allocentric spatial memory. The UAV incrementally accumulates visual observations on a fixed north-up map initialized with geographic and landmark priors, preserving evidence across viewpoint changes. A dynamic query conditioned on the instruction, current observations, UAV state, and motion history retrieves relevant memory evidence to predict a map-frame target. A deterministic controller converts this target into body-frame commands using the UAV's current pose, connecting global reasoning with egocentric execution. To reduce target fluctuations over long horizons, early goal commitment (EGC) fixes the execution target to the coordinate-wise median of the first $K$ predictions without extra trainable parameters. On Test Unseen, CityNav-VLA reduces NE by 21.2\% and improves SR and SPL by 15.8\% and 30.9\%, respectively. It also reduces NE by 6.2\% and improves SR and SPL by 5.6\% and 23.2\%, respectively, relative to the former SOTA after dataset refinement. Cumulative ablations show that navigation-map features yield the largest gains, while metric coordinates further improve endpoint accuracy. With a 4B-scale backbone, inference averages 0.68 s per decision with 8.6 GB peak memory on a single A100, presenting remarkable model efficiency.
\end{abstract}

\begin{figure}[t]
  \centering
  \includegraphics[width=0.48\textwidth]{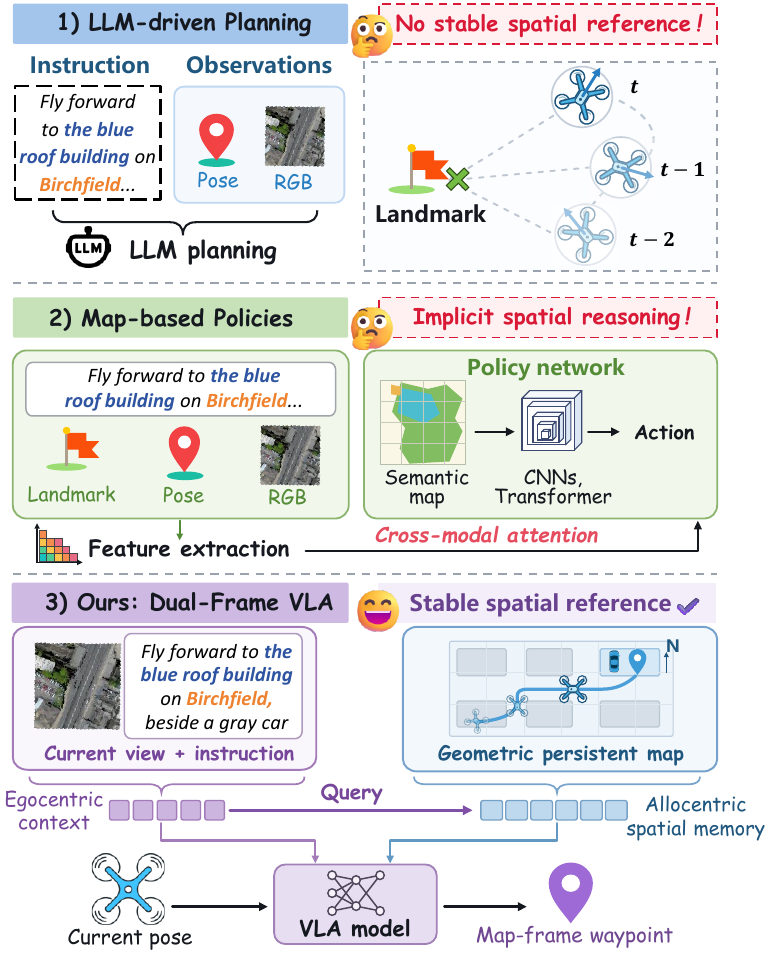}
  %\caption{Navigation architectures for language-goal aerial navigation. (1) LLM-driven planning reasons from the instruction and current egocentric observations, whose spatial references change as the UAV moves. The inset illustrates how the bearing to the same landmark varies with the UAV pose. (2) Map-based policies maintain spatial information in an explicit map and combine it with language and observations for navigation. (3) CityNav-VLA separates two complementary reference frames: dynamic egocentric interaction and persistent allocentric spatial memory. The egocentric context queries the allocentric memory to predict a map-referenced waypoint, which is transformed back into the UAV-centered body frame for execution.}
  \caption{\small Language-goal aerial navigation paradigms (\S\ref{sec:introduction}). (1) LLM-driven planning uses instructions and egocentric observations, with pose-dependent bearings to the same landmark. (2) Map-based policies fuse map features implicitly within black-box networks. (3) CityNav-VLA queries persistent allocentric spatial memory with dynamic egocentric context to predict a map-frame waypoint, then converts it into body-frame commands using the current pose.}
  \vspace{-2pt}
  \label{fig:teaser}
\end{figure}
\section{Introduction}
\label{sec:introduction}

Language-goal aerial navigation (LGAN)~\cite{lee2025citynav,fan2023aerial,liu2023aerialvln} requires a UAV to reach a destination specified by a natural-language instruction, typically describing its appearance, surrounding landmarks, and spatial relations. Because the target often lies outside the current field of view, the policy must relate local observations to spatial evidence accumulated along the flight path. Existing approaches mainly follow two paradigms: LLM-driven egocentric planning~\cite{zhang2024navid,zhou2024navgpt,cai2025flightgpt} and map-based navigation policies~\cite{zhang2025citynavagent,xu2026geonav,fan2026htnav}, as illustrated in panels 1 and 2 of Fig.~\ref{fig:teaser}. The former reasons over observations that change with the UAV's viewpoint, making it difficult to maintain consistent spatial references over long trajectories. The latter preserves environmental information in maps, yet typically fuses map features implicitly within black-box policy networks without explicit spatial reasoning over long-range targets. Recently, vision-language-action (VLA) models~\cite{rt2,openvla} have offered a unified framework for learning perception-to-action mappings. However, applying VLA models to city-scale navigation still requires a way to retain spatial evidence across changing viewpoints and use it for long-range target prediction.

The central challenge lies in connecting two complementary reference frames. In the egocentric frame, perception and actions are relative to the UAV's current position and heading, so the same landmark appears at different relative positions during flight. An allocentric map instead provides a fixed reference for organizing landmarks and accumulated observations. Yet a global map alone is insufficient: the UAV must identify the instruction-specified target and determine how to approach it from its current pose. Because this pose continuously changes, even a fixed map-frame target requires different body-frame commands. Navigation therefore requires a connection between first-person perception, global spatial reasoning, and local execution.

To this end, we propose \textbf{CityNav-VLA}, a \textbf{dual-frame VLA architecture} for city-scale aerial navigation (Fig.~\ref{fig:teaser}, panel 3). Unlike prior map-based methods that rely on implicit feature fusion (Fig.~\ref{fig:teaser}, panel 2), our dual-frame design explicitly decouples allocentric spatial memory from egocentric execution, anchoring visual evidence to metric coordinates for explicit target grounding. Inspired by findings that navigational strategies shape hippocampal spatial representations~\cite{gobbo2026navigational}, our design assigns complementary roles to these frames: allocentric memory anchors environmental evidence to fixed locations for long-range reasoning, while egocentric context supports perception and action from the current UAV pose.

To ground target prediction, we develop \textbf{persistent allocentric memory with explicit metric grounding}. A fixed north-up map is initialized with landmark priors, then accumulates newly observed RGB regions, exploration masks, and flight trajectories at corresponding map locations. Its position, orientation and scale remain fixed throughout an episode, making previously observed regions accessible even after they leave the current field of view. At each step, visual tokens from observed map imagery are aligned with navigation-map features on a shared spatial grid. Learned positional embeddings and metric coordinate encodings anchor cell to a fixed physical location, preserving visual evidence and spatial relations across viewpoint changes.

A frozen pretrained multimodal backbone encodes the instruction, current observations, and accumulated map imagery. The resulting context feature is combined with UAV state and motion history to form a dynamic navigation query. Through cross-attention, the query retrieves relevant memory evidence and predicts the destination as a map-frame waypoint. To mitigate target fluctuations over long navigation horizons, we further introduce \textbf{early goal commitment (EGC)}, a lightweight strategy that takes the coordinate-wise median of the first $K$ predictions as a fixed execution target, without introducing additional trainable parameters. A deterministic controller then transforms this target into body-frame commands based on the UAV's current position and heading, enabling adaptive egocentric execution while maintaining a stable global target.

% On CityNav Test Unseen, CityNav-VLA achieves a success rate (SR) of 26.59\% and an SPL of 25.70\% under three-dimensional evaluation, and 30.52\% SR under planar evaluation on refined CityNav. Ablations show that persistent visual history improves navigation, allocentric metric grounding strengthens endpoint localization, and EGC improves long-horizon success with unchanged policy weights. With a 4B-scale backbone, inference averages 0.68\,s per decision and uses 8.56\,GB of peak memory on a single A100 GPU.
% On CityNav Test Unseen, CityNav-VLA achieves a success rate (SR) of 26.59\% and an SPL of 25.70\% under three-dimensional evaluation, and 30.52\% SR under planar evaluation on refined CityNav. Ablations show that persistent visual history improves navigation, allocentric metric grounding strengthens endpoint localization, and EGC improves long-horizon success with unchanged policy weights. With a 4B-scale backbone, inference averages 0.68\,s per decision and uses 8.56\,GB of peak memory on one A100 GPU.

On Test Unseen, CityNav-VLA reduces NE by 21.2\% and improves SPL by 26.1\% on CityNav, with corresponding gains of 6.2\% and 23.2\% on refined CityNav, respectively, relative to the best prior results for each metric. Ablations highlight navigation-map features as the primary source of performance gains, with metric coordinates and EGC providing further improvements.

\section{Related Work}
\label{sec:related_work}

\noindent\textbf{Language-Goal Aerial Navigation.} Continuous and city-scale benchmarks~\cite{fan2023aerial,liu2023aerialvln,wang2025towards,xiao2025uav,lee2025citynav,gao2025openfly} require UAVs to ground language-defined destinations in visual appearance and landmark relations~\cite{miyanishi2023cityrefer}. Existing approaches leverage VLM guidance~\cite{spf2025,xu2026fly0}, learned policies~\cite{singer2025,discretevla2025}, and structured reasoning~\cite{cai2025flightgpt,ning2026lookasidevln} to make flight decisions based on current observations, while some further exploit anticipated observations to guide action selection and trajectory planning~\cite{zhang2026anwm,liu2026imagineuav,xu2026aerialvla}. To extend spatial context beyond the current field of view, map-based approaches incorporate spatial memory and hierarchical planning, supporting long-range localization and subgoal selection~\cite{zhang2025citynavagent,xu2026geonav,fan2026htnav}. Nevertheless, maintaining spatial context does not necessarily enable direct use of accumulated visual evidence for final target prediction. How to anchor such evidence to fixed metric locations and leverage it for target prediction remains underexplored. CityNav-VLA addresses this gap by integrating persistent allocentric memory with metric-grounded target prediction.

\begin{figure*}[t]
    \centering
    \includegraphics[width=\textwidth]{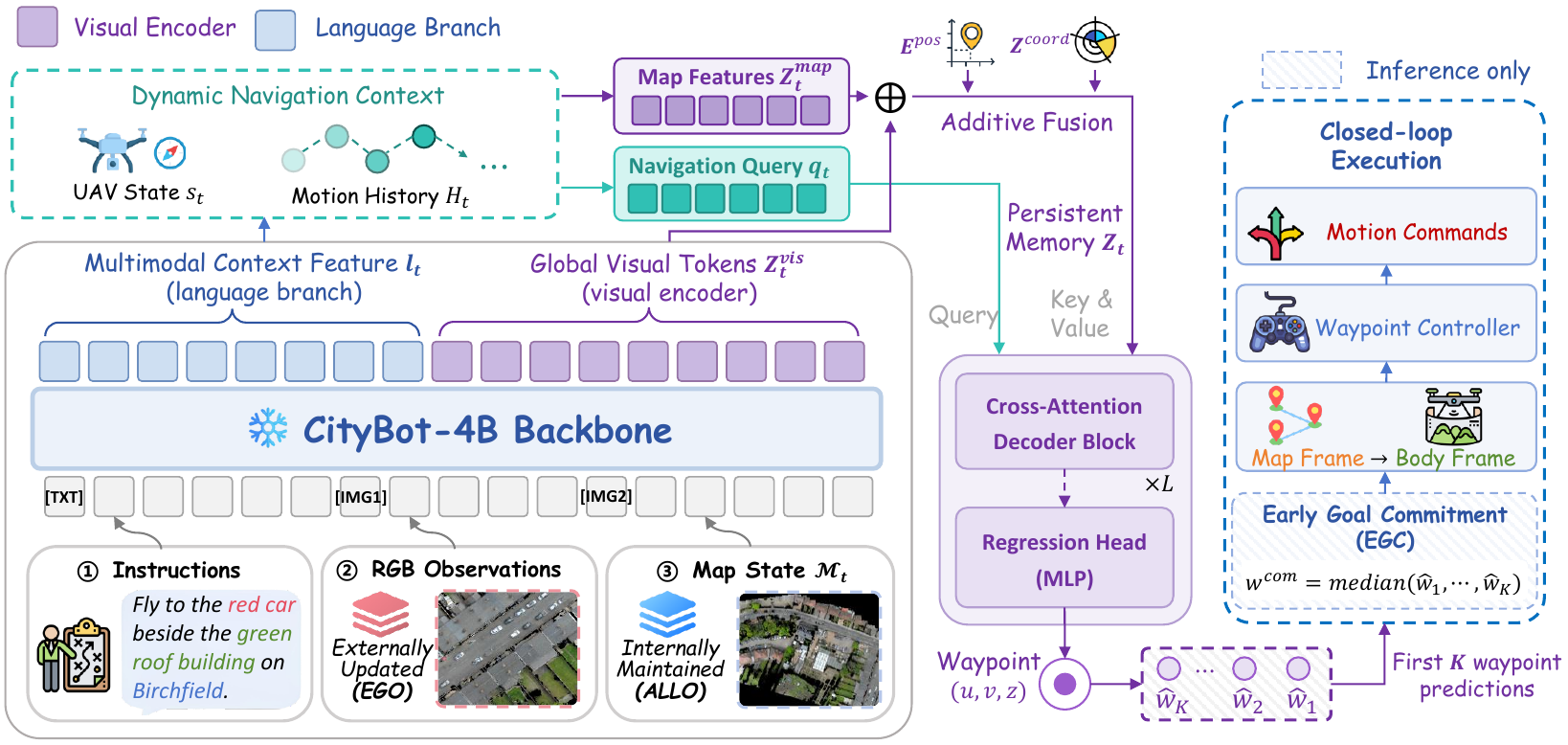}
    \caption{\small  CityNav-VLA overview (\S\ref{sec:method}). CityBot-4B encodes the instruction, local RGBs, and global map view. Dynamic context queries memory is re-encoded from $\mathcal{M}_t$ at each step to predict the final destination as a map-frame waypoint. EGC selects the execution target, whose displacement is converted into body-frame commands using the current pose.}
    \label{fig:method_overview}
    \vspace{-5pt}
\end{figure*}

\noindent\textbf{Spatial Memory and Reference Frames.} Navigation in large environments requires retaining observations beyond the current step. Spatial memory has been realized as dense language-embedded maps that fuse visual-language features with 3D reconstructions~\cite{huang2023vlmaps}, volumetric grids~\cite{liu2024ver}, annotated semantic maps ~\cite{zhang2025mapnav}, and Gaussian-language representations~\cite{zhang2026multi,xu2026map}. Learned memories compress this history into dual implicit semantic-geometric channels~\cite{zeng2025janusvln} or jointly optimize map generation with action prediction~\cite{lian2026mapdream}. A recurring choice is the reference frame. Egocentric coordinates depend on the current pose, whereas allocentric maps anchor locations independently of it, and biological navigation exploits both systems in tandem~\cite{gobbo2026navigational}. LGAN exposes the fragility: agents suffer from navigation amnesia when critical evidence lies far behind the current observation~\cite{wei2026memoryvln,lv2026himemvln}. Explicit spatial structures must therefore be not only built but made reliably retrievable and coupled to control. CityNav-VLA addresses this by accumulating exploration history in a persistent map and encoding it into spatially aligned memory tokens at each decision. Metric coordinates anchor each token to a physical map location, so a query retrieves appearance and location together.

\noindent\textbf{Spatial Representations in VLA Models.} VLA models adapt pretrained multimodal representations for action prediction~\cite{rt2,openvla}. SpatialVLA uses Ego3D position encoding for manipulation~\cite{qu2025spatialvla}. SeeR-VLA transforms point maps into a robot-centric frame to align geometry with control~\cite{lee2026see}. Action chunking and tokenization ~\cite{zhao2023act, pertsch2025fast} structure action prediction, while Qwen-VLA unifies manipulation and navigation across embodiments~\cite{qwenvla2026}. Video-based VLAs such as NaVid predict actions from current and recent visual context without persistent spatial memory~\cite{zhang2024navid}. Current-step geometric grounding thus leaves open how past observations support later goal localization. In manipulation, SOMA shows persistent multi-view memory can localize out-of-view targets~\cite{li2026soma}. City-scale navigation must combine evidence from different poses into a destination estimate after landmarks leave view. AirAnchor maintains persistent object memory for zero-shot aerial VLN~\cite{fan2026airanchor}, and Fly0 projects image-grounded targets into 3D for geometric planning~\cite{xu2026fly0}. CityNav-VLA uses instruction-conditioned egocentric context to retrieve history from persistent allocentric memory and predict a destination on the geometric consistency map.

%Geometric grounding at the current step leaves the question of how earlier observations should support later goal localization. City-scale aerial navigation requires evidence acquired at different poses to contribute to a common destination estimate after individual landmarks leave view. Fly0 projects an image-grounded target into 3D and follows it through geometric planning~\cite{xu2026fly0}. CityNav-VLA focuses on retrieving the accumulated evidence used to predict that target. Dynamic instruction-conditioned context queries persistent allocentric memory for a map-referenced navigation target, whose displacement is transformed into body-frame commands during execution. This connects long-term spatial memory to current target prediction while preserving the geometric relation between memory and control.

% \input{secv1/3_method}

\section{Method}
\label{sec:method}

\subsection{Problem Formulation}
\label{sec:problem_formulation}

Language-goal aerial navigation (LGAN) requires a UAV to reach a destination specified by a natural-language instruction $\mathcal{I}$ in a continuous urban environment. The ground-truth destination $\mathbf{g}^{\ast}$ is unknown to the UAV during navigation. At each decision step $t$, the policy uses
\begin{equation}
    \mathcal{X}_t =
    (\mathcal{I}, \mathbf{o}_t, \mathbf{s}_t,
    \mathcal{H}_t, \mathcal{M}_t),
    \label{eq:policy_input}
\end{equation}
where $\mathbf{o}_t$ denotes the current RGB observations, $\mathbf{s}_t$ represents the UAV state, including its position and heading, $\mathcal{H}_t$ records recent motion history, and $\mathcal{M}_t$ is an internally maintained map that accumulates spatial evidence over time.

Rather than predicting low-level motion commands directly, CityNav-VLA estimates the final destination in the fixed allocentric map frame $\mathcal{F}_M$:
\begin{equation}
    \hat{\mathbf{w}}_t =
    \pi_{\theta}(\mathcal{X}_t),
    \label{eq:target_prediction}
\end{equation}
where $\pi_{\theta}$ is the learned navigation policy and $\hat{\mathbf{w}}_t$ denotes the predicted destination in normalized map coordinates. The execution policy selects a target $\mathbf{w}^{\mathrm{exec}}_t$ from these predictions. A deterministic controller then converts it into a bounded motion command in the current UAV body frame $\mathcal{F}_{B_t}$:
\begin{equation}
    \mathbf{a}_t =
    \mathcal{C}\left(
    \mathcal{T}_{M \rightarrow B_t}
    (\mathbf{w}^{\mathrm{exec}}_t, \mathbf{s}_t)
    \right),
    \label{eq:body_control}
\end{equation}
where $\mathcal{T}_{M \rightarrow B_t}$ converts the selected map-frame target into a body-frame displacement, and $\mathcal{C}$ denotes the controller. After each action, the UAV updates its observations, state, motion history, and map memory. The objective is to reach the specified destination within the episode horizon.

\subsection{Method Overview}
\label{sec:method_overview}

%CityNav-VLA connects persistent allocentric memory with egocentric execution through map-frame target prediction (Fig.~\ref{fig:method_overview}). 
As shown in Fig.~\ref{fig:method_overview}, CityNav-VLA connects persistent allocentric memory with egocentric execution through map-frame target prediction.
The map provides a stable reference for target reasoning, while execution adapts to the UAV's changing position and heading. Its map state $\mathcal{M}_t$ combines geographic and landmark priors with accumulated visual evidence and flight history (Sec.~\ref{sec:map_construction}). A frozen CityBot-4B backbone, adapted from Qwen3-VL-4B for cross-view spatial reasoning~\cite{xu2026citycube}, processes the instruction, local RGB views, and the observed map imagery, termed the \emph{global map view}. The local views comprise a heading-aligned observation and a north-up crop. The backbone returns a context feature $\mathbf{l}_t$ and global visual tokens $\mathbf{Z}^{\mathrm{vis}}_t$. The latter are fused with navigation-map features, positional embeddings, and metric coordinates into memory $\mathbf{Z}_t$ (Sec.~\ref{sec:spatial_memory}).

The context, UAV state, and motion history form a query $\mathbf{q}_t$, which retrieves memory through cross-attention to predict the final destination (Sec.~\ref{sec:waypoint_head}). Training supervises this destination at every step (Sec.~\ref{sec:waypoint_training}). During inference, EGC selects a stable target from the first $K$ predictions, and a deterministic controller converts it into body-frame commands using the current pose (Sec.~\ref{sec:closed_loop}). Additional implementation details are given in Appendix~\ref{sec:additional_method_details}.

\subsection{Allocentric Map Construction}
\label{sec:map_construction}

We maintain $\mathcal{M}_t$ on a fixed north-up canvas of resolution $H\times W$ throughout each episode. CityNav's prior geographic coordinates map world locations to fixed canvas positions. Terrain height, all landmark contours, and instruction-referenced landmark masks are rasterized before navigation and remain fixed. Canvas padding and resizing preserve georeferencing, as detailed in Appendix~\ref{sec:additional_method_details}. The map therefore keeps its position, orientation, and scale as the UAV moves.

At step $t$, the UAV's position, height above ground, and heading determine a ground-view footprint and binary visibility mask $V_t\in\{0,1\}^{H\times W}$. The explored-area mask $E_t$ and accumulated RGB map $A_t$ are updated as
\begin{equation}
E_t=\bigvee_{\tau=0}^{t}V_\tau,\qquad
A_t=E_t\odot A^{\mathrm{geo}},
\label{eq:map_accumulation}
\end{equation}
where $A^{\mathrm{geo}}$ is the simulator-provided registered overhead RGB image. The operator $\bigvee$ denotes pixel-wise logical OR, while $\odot$ applies the explored mask to all RGB channels. Only explored regions are revealed; unseen regions remain masked, and past observations retain their map locations after leaving the current view. The complete map state is
\begin{equation}
\begin{aligned}
\mathcal{M}_t &= \operatorname{Concat}_{\mathrm{ch}}\bigl(
 A_t,H^{\mathrm{geo}},L^{\mathrm{all}},\\
 &\qquad L^{\mathrm{ref}},V_t,E_t,P_t\bigr)
 \in\mathbb{R}^{H\times W\times9},
\end{aligned}
\label{eq:map_state}
\end{equation}
where $H^{\mathrm{geo}}$ is terrain height, $L^{\mathrm{all}}$ and $L^{\mathrm{ref}}$ mark all and instruction-referenced landmarks, and $P_t$ records visited positions. The three RGB channels and six single-channel layers share one map grid. The accumulated RGB layer $A_t$ serves as the backbone's global map view, while the complete $\mathcal{M}_t$ is encoded into navigation-map features.

\subsection{Allocentric Memory Encoding}
\label{sec:spatial_memory}

\begin{figure*}[t]
\centering
\includegraphics[width=\textwidth]{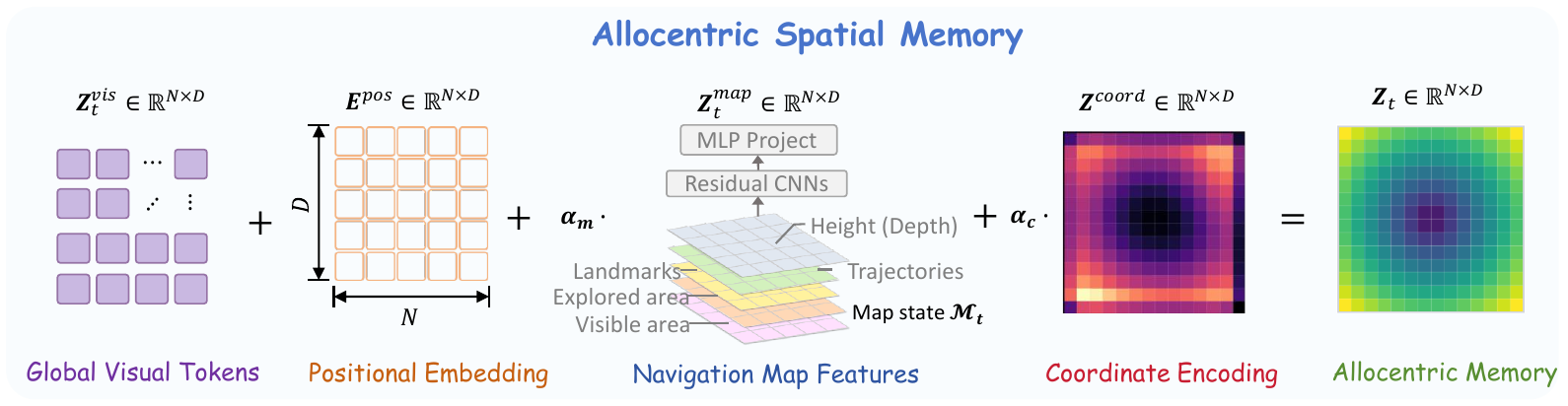}
\caption{\small Allocentric memory encoding (\S\ref{sec:spatial_memory}). Global visual tokens $\mathbf{Z}^{\mathrm{vis}}_t$ and navigation-map features $\mathbf{Z}^{\mathrm{map}}_t$ share a $14\times14$ grid and are fused with positional embeddings $\mathbf{E}^{\mathrm{pos}}$ and metric coordinate encodings $\mathbf{Z}^{\mathrm{coord}}$. Learnable scalars $\alpha_m$ and $\alpha_c$ weight the map and coordinate branches; $D=256$.}
\label{fig:spatial_memory}
\end{figure*}

At each step, we re-encode $\mathcal{M}_t$ into spatial memory $\mathbf{Z}_t\in\mathbb{R}^{N\times D}$, with $N=14\times14=196$ and $D=256$ (Fig.~\ref{fig:spatial_memory}). Each token describes a fixed map cell's appearance, navigation cues, and physical location.

The backbone's visual encoder processes the accumulated RGB map $A_t$, and its features are projected onto the shared grid to obtain $\mathbf{Z}^{\mathrm{vis}}_t\in\mathbb{R}^{N\times D}$. A convolutional encoder processes the complete nine-channel $\mathcal{M}_t$ to obtain navigation-map features $\mathbf{Z}^{\mathrm{map}}_t\in\mathbb{R}^{N\times D}$. Corresponding tokens in both branches describe the same map region. The visual branch retains observed appearance, while the map branch supplies geographic, landmark, visibility, and trajectory cues.

A learned positional embedding $\mathbf{E}^{\mathrm{pos}}\in\mathbb{R}^{N\times D}$ identifies grid positions but does not explicitly encode physical distances. We therefore construct metric coordinates from each cell's physical center $\mathbf{p}_i=(x_i,y_i)$ and map center $\bar{\mathbf{p}}=(\bar{x},\bar{y})$ in $\mathcal{F}_M$:
\begin{equation}
\mathbf{c}_i=\frac{1}{s_c}\left[
x_i-\bar{x},\;y_i-\bar{y},\;
\|\mathbf{p}_i-\bar{\mathbf{p}}\|_2
\right],
\label{eq:metric_coordinate_encoding}
\end{equation}
The first two components encode planar map-axis offsets, and the third encodes radial distance from the map center. We use $s_c=400\,\mathrm{m}$ as a fixed normalization hyperparameter. A convolutional encoder maps this coordinate field to $\mathbf{Z}^{\mathrm{coord}}\in\mathbb{R}^{N\times D}$; the coordinates remain fixed throughout the episode. The four spatially aligned components are fused additively:
\begin{equation}
\mathbf{Z}_t
=
\mathbf{Z}^{\mathrm{vis}}_t
+
\mathbf{E}^{\mathrm{pos}}
+
\alpha_m\mathbf{Z}^{\mathrm{map}}_t
+
\alpha_c\mathbf{Z}^{\mathrm{coord}},
\label{eq:spatial_fusion}
\end{equation}
where $\alpha_m$ and $\alpha_c$ are learnable scalars for navigation-map and metric coordinate features. The resulting memory associates accumulated evidence with fixed physical locations.

\subsection{Navigation Query and Target Grounding}
\label{sec:waypoint_head}

The dynamic query selects map evidence relevant to the instruction and the UAV's current situation. An MLP encodes the state $\mathbf{s}_t$, and a GRU encodes the recent motion history $\mathcal{H}_t$. Their outputs are projected to the memory dimension and combined with $\mathbf{l}_t$ to form a single query token:
\begin{equation}
\mathbf{q}_t
=
\operatorname{LN}
\left(
\mathbf{W}_{l}\mathbf{l}_t
+
f_s(\mathbf{s}_t)
+
f_h(\mathcal{H}_t)
\right),
\label{eq:navigation_query}
\end{equation}
where $\mathbf{W}_l$ is a learned projection, $f_s$ and $f_h$ are the state and history encoders, and $\operatorname{LN}$ is layer normalization. The query retrieves spatial evidence through $L$ cross-attention decoder blocks, with $\mathbf{Z}_t$ providing the keys and values at every block:
\begin{equation}
\widetilde{\mathbf{q}}_t=\operatorname{Dec}_{L}(\mathbf{q}_t,\mathbf{Z}_t).
\label{eq:cross_attention}
\end{equation}
Each block contains cross-attention and a feed-forward layer with residual connections. An MLP regression head maps $\widetilde{\mathbf{q}}_t$ to three raw values $(r_u,r_v,r_z)$ and produces
\begin{equation}
\hat{\mathbf{w}}_t
=
\left(
\sigma(r_u),\;\sigma(r_v),\;\tanh(r_z)
\right),
\label{eq:waypoint_head}
\end{equation}
where $\sigma$ is the sigmoid function that bounds the planar coordinates $(\hat{u}_t,\hat{v}_t)$ to the normalized map domain. The vertical output $\hat{z}_t$ is the height relative to the map's ground reference, normalized by $s_z=100\,\mathrm{m}$ and bounded by $\tanh$. The output remains in the map frame until execution.

\subsection{Training Objective}
\label{sec:waypoint_training}

The navigation target $\hat{\mathbf{w}}_t=(\hat{u}_t,\hat{v}_t,\hat{z}_t)$ estimates the episode's final destination at every step. 
% The normalized supervision $(u^{*},v^{*},z^{*})$ is fixed throughout the trajectory, so each prediction grounds the same destination using the evidence available at that step. 
The normalized supervision $(u^{},v^{},z^{*})$ is fixed along the trajectory, so each prediction grounds the same destination using the evidence available at that step.
To account for different map extents, we convert prediction errors to meters using the map-specific planar scale $\mathbf{m}=(m_u,m_v)$ and vertical scale $s_z$:
\[
\begin{aligned}
\mathbf{e}^{xy}_t&=
[(\hat{u}_t-u^{*}),(\hat{v}_t-v^{*})]\odot\mathbf{m},\\
e^z_t&=s_z(\hat{z}_t-z^{*}),
\end{aligned}
\]
where $\odot$ denotes element-wise multiplication. We minimize the weighted Smooth-$L_1$ target-coordinate loss:
\begin{equation}
\mathcal{L}_{\mathrm{goal}}
=
\frac{1}{2}
\sum_{d\in\{x,y\}}
\rho\!\left(\frac{e_t^{d}}{\delta}\right)
+
\lambda_z
\rho\!\left(\frac{e_t^{z}}{\delta}\right),
\label{eq:goal_loss}
\end{equation}
where $\rho$ is the Smooth-$L_1$ penalty, $\delta=20\,\mathrm{m}$ sets the error scale, and $\lambda_z=0.25$ weights the vertical component. The loss is averaged over training samples.

CityBot-4B remains frozen, while the input projections, state and history encoders, map and coordinate encoders, positional embeddings, fusion weights, decoder, and regression head are trained jointly with $\mathcal{L}_{\mathrm{goal}}$.

\subsection{Target Execution Policy}
\label{sec:closed_loop}

\noindent\textbf{Early goal commitment (EGC).} EGC stabilizes the destination estimate while the UAV's observations continue to change. During the first $K$ decisions, the controller follows the raw predictions $\hat{\mathbf{w}}_1,\ldots,\hat{\mathbf{w}}_K$. It then uses their coordinate-wise median as the committed target in the remaining steps:
\begin{equation}
\mathbf{w}^{\mathrm{com}}
=
\operatorname{median}
\left(
\hat{\mathbf{w}}_1,\ldots,\hat{\mathbf{w}}_K
\right).
\label{eq:early_commitment}
\end{equation}
From decision $K+1$ onward, $\mathbf{w}^{\mathrm{com}}$ remains the execution target. EGC operates only during inference and adds no trainable parameters; when disabled, the controller follows the current prediction at every decision.

\noindent\textbf{Map-to-body control.} The selected target is converted from normalized map coordinates to a physical position using the map georeferencing transform. Let $(\Delta x_t,\Delta y_t)$ denote its planar displacement from the current UAV position in $\mathcal{F}_M$. Given heading $\psi_t$, the corresponding body-frame displacement is
\begin{equation}
\begin{bmatrix}
\Delta p^{\mathrm{fwd}}_t \\
\Delta p^{\mathrm{left}}_t
\end{bmatrix}
=
\mathbf{R}(-\psi_t)
\begin{bmatrix}
\Delta x_t \\
\Delta y_t
\end{bmatrix},
\label{eq:body_transform}
\end{equation}
where $\mathbf{R}(-\psi_t)$ is the planar rotation from the map frame to the body frame. A deterministic controller converts the forward, lateral, and vertical displacements into bounded motion commands. Observations and map memory continue to update, while the controller recomputes body-frame displacements from the current pose at every decision, including after commitment to the fixed target. Controller settings are given in Appendix~\ref{sec:experimental_details}.

\section{Experiments}

We evaluate navigation performance, then examine how spatial memory and target commitment contribute to successful arrival. The ablations distinguish the information retained in memory from the rule used to execute its predictions.

\subsection{Experimental Setup}
\label{sec:experimental_setup}

\begin{table*}[t]
\centering
\caption{\small Navigation performance on CityNav and refined CityNav (\S\ref{sec:main_results}). CityNav-VLA uses three-dimensional distances on CityNav and planar distances on refined CityNav. Published baselines retain their original evaluation settings. HETT$^*$ denotes the refined dataset result. NE is in meters; SR, OSR and SPL are percentages. Bold/underline indicate the best/second-best reported model results within each dataset.}
\label{tab:results}
\centering
\small
\setlength{\tabcolsep}{6pt}
\renewcommand{\arraystretch}{1.1}
\resizebox{\linewidth}{!}{
% \begin{tabular*}{\linewidth}{@{\extracolsep{\fill}}rrrrrrrrrrrrr@{}}
\begin{tabular}{rrrrrrrrrrrrrr}
\toprule
\multirow{2}{*}{Method} & \multicolumn{4}{c}{Validation Seen} & \multicolumn{4}{c}{Validation Unseen} & \multicolumn{4}{c}{\textbf{Test Unseen}} \\
\cmidrule(lr){2-5}\cmidrule(lr){6-9}\cmidrule(lr){10-13}
 & NE$\downarrow$ & SR$\uparrow$ & OSR$\uparrow$ & SPL$\uparrow$ & NE$\downarrow$ & SR$\uparrow$ & OSR$\uparrow$ & SPL$\uparrow$ & NE$\downarrow$ & SR$\uparrow$ & OSR$\uparrow$ & SPL$\uparrow$ \\
\midrule
\multicolumn{13}{@{}c}{\textit{CityNav}} \\
\midrule
% Random~\cite{lee2025citynav} & 222.30 & 0.00 & 1.15 & 0.00 & 223.00 & 0.00 & 0.90 & 0.00 & 208.80 & 0.00 & 1.44 & 0.00 \\
% Seq2Seq+GSM~\cite{lee2025citynav} & 58.50 & 8.43 & 17.31 & 7.28 & 78.60 & 5.13 & 10.90 & 4.65 & 98.10 & 3.81 & 13.92 & 2.79 \\
% CMA+GSM~\cite{lee2025citynav} & 68.00 & 6.25 & 13.28 & 5.40 & 75.90 & 4.38 & 9.29 & 3.90 & 94.60 & 4.68 & 12.01 & 4.05 \\
% MGP+LLaVA (HD)~\cite{lee2025citynav} & 59.70 & 8.69 & 35.51 & 8.28 & 75.10 & 5.84 & 22.19 & 5.56 & 93.80 & 6.38 & 26.04 & 6.08 \\
% AerialVLN+GSM~\cite{liu2023aerialvln,lee2025citynav} & 56.60 & 10.16 & 22.20 & 7.89 & 72.70 & 6.35 & 15.24 & 5.06 & 85.10 & 6.72 & 18.21 & 5.16 \\
FLAME~\cite{xu2025flame} & 59.30 & 19.44 & 29.49 & 17.21 & 69.60 & 14.83 & 28.31 & 13.85 & 80.40 & 20.81 & 32.35 & 19.15 \\
FlightGPT~\cite{cai2025flightgpt} & 66.10 & 17.57 & 30.26 & 15.78 & 68.10 & 14.69 & 29.33 & 13.24 & 76.20 & 21.20 & 35.38 & 19.24 \\
GeoNav~\cite{xu2026geonav} & 58.60 & 21.80 & 39.98 & 12.37 & 64.10 & \underline{16.99} & \underline{32.41} & 9.96 & 73.50 & \underline{25.91} & \textbf{41.68} & 16.09 \\
DroneNav~\cite{liu2026dronenav} & 54.30 & 18.94 & 36.65 & 17.37 & 63.50 & 15.06 & 25.81 & 13.27 & 74.90 & 22.47 & 30.22 & 20.38 \\
HTNav~\cite{fan2026htnav} & 47.20 & \underline{28.30} & \underline{47.92} & \underline{25.07} & \textbf{59.80} & 15.85 & \textbf{34.43} & 13.55 & 68.50 & 22.23 & \underline{41.02} & 19.63 \\
HETT~\cite{ding2026hett} & 45.20 & 25.16 & \textbf{48.40} & 23.01 & \underline{62.10} & \textbf{17.48} & 25.09 & 14.46 & 72.90 & 22.97 & 39.30 & 17.01 \\
Human~\cite{lee2025citynav} & 9.10 & 89.31 & 96.40 & 60.17 & 9.40 & 88.39 & 95.54 & 62.66 & 9.80 & 87.86 & 95.29 & 57.04 \\
\midrule
CityNav-VLA w/o EGC & \underline{43.10} & 26.14 & 31.14 & 24.72 & 69.56 & 15.78 & 22.29 & \underline{14.83} & \underline{58.35} & 22.16 & 28.04 & \underline{21.15} \\
\rowcolor[gray]{0.93}
\textbf{CityNav-VLA} & \textbf{40.08} & \textbf{28.38} & 32.31 & \textbf{27.20} & 65.94 & 15.99 & 21.59 & \textbf{15.26} & \textbf{54.01} & \textbf{26.59} & 30.95 & \textbf{25.70} \\
\midrule
\multicolumn{13}{@{}c}{\textit{refined CityNav}} \\
\midrule
HETT$^*$~\cite{ding2026hett} & \underline{37.20} & \textbf{31.09} & \textbf{51.86} & 25.76 & \textbf{51.30} & \textbf{19.10} & \textbf{34.78} & \textbf{16.70} & 40.40 & \underline{28.90} & \textbf{49.56} & 23.79 \\
\midrule
CityNav-VLA w/o EGC & 38.62 & 28.14 & \underline{43.08} & \underline{26.02} & 55.78 & 14.76 & \underline{31.15} & 13.61 & \underline{40.14} & 26.66 & 42.40 & \underline{24.84} \\
\rowcolor[gray]{0.93}
\textbf{CityNav-VLA} & \textbf{36.48} & \underline{29.96} & 41.74 & \textbf{28.76} & \underline{51.95} & \underline{16.94} & 30.96 & \underline{16.18} & \textbf{37.91} & \textbf{30.52} & \underline{43.08} & \textbf{29.32} \\
\bottomrule
\end{tabular}}
\vspace{-5pt}
\end{table*}

\noindent\textbf{Datasets and Metrics.} We evaluate on CityNav~\cite{lee2025citynav} and refined CityNav, released with HETT~\cite{ding2026hett}. Our CityNav results use three-dimensional distances; refined CityNav results use planar distances, following HETT's evaluation. Navigation error (NE) measures final distance to the target, success rate (SR) measures arrival within 20 m, and oracle success rate (OSR) records whether the trajectory ever enters this region. 
%SPL weights successful arrival by path efficiency, with path length and initial distance measured in the same coordinate space. 
SPL weights success by path efficiency, with path length and initial distance in the same coordinate space. Full-split sizes and metric definitions appear in Appendix~\ref{sec:experimental_details}.
% Full-split sizes and metric definitions appear in Appendix~\ref{sec:experimental_details}.

\noindent\textbf{Execution.} Each episode runs for 120 decisions, with arrival assessed at the final position. The controller converts each map-frame target into simultaneous forward, lateral and vertical displacements, bounded by 4, 4 and 2 m, respectively, while keeping heading fixed. CityNav's atomic actions instead separate 5 m forward moves, 2 m vertical moves, $30^\circ$ turns and STOP. Table~\ref{tab:action_protocol} gives the action mappings and movement budgets. All our ablations share the same controller and horizon; published baselines retain their original metric and execution settings.

\noindent\textbf{Training and Comparisons.} We freeze CityBot-4B~\cite{xu2026citycube} and train separate navigation policies on the 22,002 CityNav and 21,878 refined CityNav Training Seen episodes. Each policy uses its dataset's descriptions and landmark annotations. The main checkpoints are selected on Validation Seen and held fixed across evaluation splits and EGC settings. Representation ablations compare checkpoints after 10,000 updates under the same learning-rate schedule. The visual-history diagnostic changes only the accumulated RGB input to the main policy. Training settings, sample selection and statistical procedures are given in Appendix~\ref{sec:experimental_details}.

We retain $K=24$ for EGC, chosen previously on refined CityNav Validation Seen. Sec~\ref{sec:ablation_commitment} evaluates the resulting execution policy, and Appendix~\ref{sec:commitment_step_sensitivity} examines its sensitivity on CityNav Validation Seen.

\begin{table*}[t]
\centering
\caption{\small EGC improves endpoint success across Test Unseen difficulty groups (\S\ref{sec:main_results}). Each dataset panel partitions the corresponding trajectories in Table~\ref{tab:results}; Overall is the episode-weighted mean. CityNav scores are three-dimensional and refined CityNav scores are planar. NE is in meters; other metrics are percentages. Subscripts denote changes with EGC relative to without EGC. Blue/red indicate improvements/regressions, respectively. Bold/underline indicate the better/second-best result in each pair.}
\label{tab:difficult}
\small
\setlength{\tabcolsep}{3pt}
\renewcommand{\arraystretch}{1.18}

\resizebox{\linewidth}{!}{%
\begin{tabular}{@{} l l r *{8}{c} @{}}
\toprule
\multirow{2}{*}{Dataset} & \multirow{2}{*}{Difficulty}
& \multirow{2}{*}{Episodes}
& \multicolumn{2}{c}{NE$\downarrow$}
& \multicolumn{2}{c}{SR$\uparrow$}
& \multicolumn{2}{c}{OSR$\uparrow$}
& \multicolumn{2}{c}{SPL$\uparrow$} \\
\cmidrule(lr){4-5}
\cmidrule(lr){6-7}
\cmidrule(lr){8-9}
\cmidrule(lr){10-11}
& & & w/o & w/ & w/o & w/ & w/o & w/ & w/o & w/ \\
\midrule

\multirow{4}{*}{\textbf{CityNav}} 
& Easy & 2,096
& \underline{53.76} & \egcbetter{48.30}{-5.46}
& \underline{24.38} & \egcbetter{29.29}{+4.91}
& \underline{29.77} & \egcbetter{31.39}{+1.62}
& \underline{22.86} & \egcbetter{28.12}{+5.26} \\
& Medium & 1,686
& \underline{58.51} & \egcbetter{54.79}{-3.72}
& \underline{23.19} & \egcbetter{25.50}{+2.31}
& \underline{29.30} & \egcbetter{31.02}{+1.72}
& \underline{22.42} & \egcbetter{24.82}{+2.40} \\
& Hard & 1,529
& \underline{64.47} & \egcbetter{60.98}{-3.49}
& \underline{17.99} & \egcbetter{24.07}{+6.08}
& \underline{24.26} & \egcbetter{30.28}{+6.02}
& \underline{17.40} & \egcbetter{23.33}{+5.93} \\
\cmidrule(l){2-11}
& \textbf{Overall} & \textbf{5,311}
& \underline{58.35} & \egcbetter{54.01}{-4.34}
& \underline{22.16} & \egcbetter{26.59}{+4.43}
& \underline{28.04} & \egcbetter{30.95}{+2.91}
& \underline{21.15} & \egcbetter{25.70}{+4.55} \\

\midrule
\multirow{4}{*}{\textbf{\makecell[l]{Refined\\CityNav}}} 
& Easy & 2,081
& \underline{37.00} & \egcbetter{33.41}{-3.59}
& \underline{27.73} & \egcbetter{33.97}{+6.24}
& \underline{51.61} & \egcbetter{51.75}{+0.14}
& \underline{24.46} & \egcbetter{32.10}{+7.64} \\
& Medium & 1,678
& \underline{40.31} & \egcbetter{38.94}{-1.37}
& \underline{27.35} & \egcbetter{29.80}{+2.45}
& \textbf{40.58} & \egcworse{40.11}{-0.47}
& \underline{26.35} & \egcbetter{28.97}{+2.62} \\
& Hard & 1,522
& \underline{44.24} & \egcbetter{42.92}{-1.32}
& \underline{24.44} & \egcbetter{26.61}{+2.17}
& \underline{31.80} & \egcbetter{34.49}{+2.69}
& \underline{23.70} & \egcbetter{25.90}{+2.20} \\
\cmidrule(l){2-11}
& \textbf{Overall} & \textbf{5,281}
& \underline{40.14} & \egcbetter{37.91}{-2.23}
& \underline{26.66} & \egcbetter{30.52}{+3.86}
& \underline{42.40} & \egcbetter{43.08}{+0.68}
& \underline{24.84} & \egcbetter{29.32}{+4.48} \\

\bottomrule
\end{tabular}%
}
\end{table*}

\subsection{Main Results}
\label{sec:main_results}

% \input{Tables/main_results}

% \noindent\textbf{Quantitative Results on (refined) CityNav.} 
\noindent\textbf{Quantitative Results on CityNav and Its Refined Version.}
CityNav-VLA achieves the lowest Test Unseen endpoint error and the highest SR and SPL among the autonomous methods listed in Table~\ref{tab:results}. It reaches an SR of $26.59\%$ and an SPL of $25.70\%$, with NE of $54.01\,\mathrm{m}$. Its OSR of $30.95\%$ is below several baselines, so the advantage concerns final arrival and path efficiency rather than trajectory coverage alone. Validation Seen yields an SR of $28.38\%$, while Validation Unseen remains more difficult, with an SR of $15.99\%$. On refined CityNav, CityNav-VLA obtains Test Unseen SR of $30.52\%$ and SPL of $29.32\%$. Relative to HETT$^*$, these are gains of $1.62$ and $5.53$ percentage points. The matched comparisons with and without EGC in Sec.~\ref{sec:ablation_commitment} isolate how execution contributes to these results.

\noindent\textbf{Performance across Difficulty Levels.}
%\label{sec:difficulty}
Table~\ref{tab:difficult} examines the official Easy, Medium, and Hard groups. Each reported setting partitions the corresponding Test Unseen trajectories in Table~\ref{tab:results}. Overall scores are the episode-weighted averages of the three groups. EGC improves final arrival in all three difficulty groups. Its SR gains are $4.91$, $2.31$ and $6.08$ percentage points on Easy, Medium and Hard, respectively, with corresponding improvements in NE and SPL. The largest success gain occurs on Hard, although this group still has the highest endpoint error at $60.98\,\mathrm{m}$. Refined CityNav shows the same direction of improvement in NE, SR and SPL across all three groups.

% \input{Tables/spatial_memory}

% \begin{table}[t]
% \centering
% \caption{\small Navigation-map features provide the largest gain in the cumulative spatial-memory ablation (\S\ref{sec:ablation_spatial_memory}). Results use all 2,697 refined CityNav Validation Unseen episodes, seed 42 and 10K checkpoints without EGC. $V$, $M$, $P$ and $C$ denote global visual tokens, navigation-map features, learned positional embeddings and metric coordinates. Scores are planar; NE is in meters and other metrics are percentages.}
% \label{tab:spatial-memory-components}
% \centering
% \small
% \setlength{\tabcolsep}{1.5pt}
% \renewcommand{\arraystretch}{1.1}
% \resizebox{\linewidth}{!}{
% % \begin{tabular}{\linewidth}{@{\extracolsep{\fill}}rccccrrrr@{}}
% % \begin{tabular}{\linewidth}{@{\extracolsep{\fill}}rccccrrrr@{}}
% \begin{tabular}{rccccrrrrr}
% \toprule
% Variant & $V$ & $M$ & $P$ & $C$ & NE$\downarrow$ & SR$\uparrow$ & OSR$\uparrow$ & SPL$\uparrow$ \\
% \midrule
% F0 & $\checkmark$ &  &  &  & 207.34 & 0.82 & 7.16 & 0.66 \\
% F1 & $\checkmark$ & $\checkmark$ &  &  & 58.81 & 12.27 & 25.14 & 11.28 \\
% F2 & $\checkmark$ & $\checkmark$ & $\checkmark$ &  & 57.01 & 14.31 & 29.40 & 13.22 \\
% F3 & $\checkmark$ & $\checkmark$ & $\checkmark$ & $\checkmark$ & 52.42 & 15.76 & 28.03 & 14.68 \\
% \bottomrule
% \end{tabular}}
% \end{table}

\begin{table}[t]
\centering
\caption{\small Navigation-map features provide the largest gain in the cumulative spatial-memory ablation (\S\ref{sec:ablation_spatial_memory}). Results use all 2,697 refined CityNav Validation Unseen episodes, seed 42 and 10K checkpoints without EGC. $V$, $M$, $P$ and $C$ denote global visual tokens, navigation-map features, learned positional embeddings and metric coordinates. Scores are based on planar.
% Bold/underline indicate the best/second-best results.
}
\label{tab:spatial-memory-components}
\small
\setlength{\tabcolsep}{4pt}
\renewcommand{\arraystretch}{1.1}
\resizebox{\linewidth}{!}{
\begin{tabular}{rccccrrrr}
\toprule
Variant & $V$ & $M$ & $P$ & $C$ & NE$\downarrow$ & SR$\uparrow$ & OSR$\uparrow$ & SPL$\uparrow$ \\
\midrule
F0 & $\checkmark$ & & & & 207.34 & 0.82 & 7.16 & 0.66 \\
F1 & $\checkmark$ & $\checkmark$ & & & 58.81 & 12.27 & 25.14 & 11.28 \\
F2 & $\checkmark$ & $\checkmark$ & $\checkmark$ & & \underline{57.01} & \underline{14.31} & \textbf{29.40} & \underline{13.22} \\
\rowcolor[gray]{0.93}
\textbf{F3} & $\checkmark$ & $\checkmark$ & $\checkmark$ & $\checkmark$ & \textbf{52.42} & \textbf{15.76} & \underline{28.03} & \textbf{14.68} \\
\bottomrule
\end{tabular}}
\end{table}

\begin{table}[t]
\centering
\caption{\small Accumulated visual evidence improves navigation (\S\ref{sec:ablation_spatial_memory}). Both conditions use the same policy on 500 Validation Seen episodes with EGC disabled. Only the RGB visibility mask changes; state history, exploration cues and landmarks are preserved. Scores are three-dimensional. NE is in meters and other metrics are percentages; gains are in meters or percentage points. Bold/underline indicate the best/second-best results.}
\label{tab:visual-history}
\small
\setlength{\tabcolsep}{5.5pt}
\renewcommand{\arraystretch}{1.1}
\resizebox{\linewidth}{!}{
\begin{tabular}{rrrrr}
\toprule
Map RGB evidence & NE$\downarrow$ & SR$\uparrow$ & OSR$\uparrow$ & SPL$\uparrow$ \\
\midrule
Current view only & \underline{47.13} & \underline{18.80} & \underline{24.40} & \underline{17.38} \\
\rowcolor[gray]{0.93}
\textbf{Accumulated views} & \textbf{43.62} & \textbf{24.20} & \textbf{27.80} & \textbf{22.76} \\
\midrule
Gain & -3.52 & +5.40 & +3.40 & +5.38 \\
\bottomrule
\end{tabular}}
\end{table}

\subsection{Spatial Memory Components}
\label{sec:ablation_spatial_memory}

\noindent\textbf{Cumulative Memory Components.} Starting from global visual tokens $V$, we successively add navigation-map features $M$, learned positional embeddings $P$, and allocentric metric coordinates $C$ to the spatial memory. The navigation query receives the same instruction-conditioned context, UAV state, and motion history in every variant. All variants share a 30K-step training schedule with seed 42. We evaluate their 10K-step checkpoints on all 2,697 Validation Unseen episodes of refined CityNav with EGC disabled. F3 uses the complete $V+M+P+C$ memory under this common protocol. Navigation-map features provide the largest improvement in endpoint localization, reducing NE by $148.52\,\mathrm{m}$ (Table~\ref{tab:spatial-memory-components}). Learned positional embeddings then improve all four metrics. Metric coordinates reduce NE by a further $4.58\,\mathrm{m}$, while their SR and SPL gains are not statistically significant. This progression separates the contribution of explicit map evidence from that of its geometric encoding. The latter contributes most clearly to endpoint accuracy; Appendix~\ref{sec:spatial_reference} compares the coordinate representations directly.

\noindent\textbf{Accumulated Visual Evidence.}
% \label{sec:visual_history}
Retaining previously observed RGB regions helps the policy locate its destination as the current view changes. To isolate this contribution, we replace the accumulated RGB visibility mask with the current-view footprint on 500 Validation Seen episodes, sampled across scenes and initial-distance ranges. The intervention affects the global map view, its local crop and the RGB channels of the navigation map. Landmark cues, exploration and trajectory channels, motion history and all model weights remain fixed. Both conditions disable EGC so that the target can respond to changing observations throughout the episode. Removing accumulated RGB reduces SR from $24.20\%$ to $18.80\%$ ($p=0.0067$) and increases NE by $3.52\,\mathrm{m}$ (Table~\ref{tab:visual-history}). SPL also decreases by $5.38$ percentage points. These changes show that previously observed map regions contribute to target prediction even when the policy retains pose, motion history and explicit landmark information. Persistent memory therefore provides useful visual evidence beyond the UAV's immediate footprint.

\subsection{Early Goal Commitment}
\label{sec:ablation_commitment}

\begin{table}[t]
\centering
\caption{\small EGC improves final arrival with a fixed navigation policy (\S\ref{sec:ablation_commitment}). Entries are EGC minus no EGC: meters for NE and percentage points for the other metrics. CityNav uses three-dimensional distances and refined CityNav uses planar distances. Each row compares the same checkpoint on the same episodes. Test Unseen gains are highlighted.}
\label{tab:ablation_commitment}
\small
\setlength{\tabcolsep}{6pt}
\renewcommand{\arraystretch}{1.1}
\resizebox{\linewidth}{!}{
\begin{tabular}{rrrrr}
\toprule
Split & $\Delta$NE$\downarrow$ & $\Delta$SR$\uparrow$ & $\Delta$OSR$\uparrow$ & $\Delta$SPL$\uparrow$ \\
\midrule
\multicolumn{5}{@{}c}{\textit{CityNav}} \\
\midrule
Validation Seen & -3.02 & +2.24 & +1.16 & +2.48 \\
Validation Unseen & -3.62 & +0.21 & -0.71 & +0.43 \\
\rowcolor[gray]{0.93}
\textbf{Test Unseen} & \textbf{-4.34} & \textbf{+4.42} & \textbf{+2.92} & \textbf{+4.55} \\
\midrule
\multicolumn{5}{@{}c}{\textit{refined CityNav}} \\
\midrule
Validation Seen & -2.14 & +1.82 & -1.34 & +2.74 \\
Validation Unseen & -3.83 & +2.19 & -0.19 & +2.57 \\
\rowcolor[gray]{0.93}
\textbf{Test Unseen} & \textbf{-2.23} & \textbf{+3.86} & \textbf{+0.68} & \textbf{+4.48} \\
\bottomrule
\end{tabular}}
\end{table}

Keeping the execution target fixed improves successful arrival without changing the learned policy. We compare EGC against continuous target updates using the same checkpoint and episodes. Table~\ref{tab:ablation_commitment} reports the split-wise comparisons on CityNav and refined CityNav.

On Test Unseen, EGC raises SR from $22.16\%$ to $26.59\%$ and SPL from $21.15\%$ to $25.70\%$. The success gain is significant ($p<0.001$), with 496 episodes becoming successful and 261 losing success. NE decreases by $4.34\,\mathrm{m}$, while OSR increases by $2.92$ percentage points. Commitment thus improves both final arrival and target-region coverage on this split. Validation Seen shows an SR gain of $2.24$ percentage points ($p=0.0065$), with lower NE and higher SPL.

On Validation Unseen, EGC reduces NE by $3.62\,\mathrm{m}$, but SR increases by only $0.21$ percentage points ($p=0.784$). SPL rises by $0.43$ points and OSR falls by $0.71$ points, with neither difference statistically significant. Commitment brings the final position closer to the target on this split without substantially changing how often the UAV arrives within the success radius.

The refined CityNav comparison also improves NE, SR and SPL on all three splits. Its OSR changes are small and not statistically significant. Across these evaluations, commitment consistently reduces endpoint error, while its effect on successful arrival varies by split. A fixed target lets the controller continue approaching an early estimate instead of following subsequent changes in the predicted destination. \ref{sec:commitment_step_sensitivity} examines how long the policy should gather predictions before committing.

\noindent\textbf{Qualitative Results on Refined CityNav.} Appendix~\ref{sec:qualitative} presents navigation examples and failure cases comparing execution with and without EGC.

\subsection{Model Efficiency}
\label{sec:efficiency}

\begin{table}[t]
\centering
\caption{\small Model size and inference efficiency of MLLM-based navigation methods (\S\ref{sec:efficiency}). Published measurements are taken from DroneNav~\cite{liu2026dronenav} under its reported hardware and timing settings. Our measurements use an NVIDIA A100-SXM4-80GB GPU. 
% Bold/underline indicate the lowest/second-lowest reported values.
}
\label{tab:efficiency}
\small
\setlength{\tabcolsep}{6pt}
\renewcommand{\arraystretch}{1.1}
\resizebox{\linewidth}{!}{
\begin{tabular}{rrrr}
\toprule
Method & \shortstack{Params$\downarrow$\\(B)} & \shortstack{GPU Mem$\downarrow$\\(GB)} & \shortstack{Latency$\downarrow$\\(ms)} \\
\midrule
GSM~\cite{lee2025citynav} & 34.96 & 75.3 & 4832.6 \\
FlightGPT~\cite{cai2025flightgpt} & 8.29 & 19.6 & 8655.9 \\
DroneNav~\cite{liu2026dronenav} & \underline{7.96} & \underline{16.4} & \underline{4254.7} \\
\midrule
\rowcolor[gray]{0.93}
\textbf{CityNav-VLA (Ours)} & \textbf{4.46} & \textbf{8.56} & \textbf{679.7} \\
\bottomrule
\end{tabular}}
\end{table}

CityNav-VLA learns navigation through a small set of modules attached to a frozen multimodal backbone. Only $4.67$\,M of its $4.46$\,B parameters are trainable. The metric coordinate branch adds $0.33$\,M parameters, so the improvement in endpoint accuracy observed in Sec.~\ref{sec:ablation_spatial_memory} comes with a small increase in model size. EGC further stabilizes execution without adding trainable parameters.

At inference, the policy predicts a waypoint from a single multimodal forward pass. On an NVIDIA A100-SXM4-80GB GPU, the backbone and waypoint head together take $679.7$ ms per decision on average and use $8.56$ GB of peak memory. Table~\ref{tab:efficiency} places these measurements alongside published costs of MLLM-based navigation systems; the hardware and timing scope of each source evaluation apply.

% \begin{table}[t]
% \centering
% \caption{Model size and inference efficiency of MLLM-based navigation methods.
% Published measurements are taken from DroneNav~\cite{liu2026dronenav} under its reported hardware and timing settings. Our measurements use an NVIDIA A100-SXM4-80GB GPU.}
% \label{tab:efficiency}
% \small
% \setlength{\tabcolsep}{3pt}
% \begin{tabular}{@{}lrrr@{}}
% \toprule
% Method & \shortstack{Params\\(B)} & \shortstack{GPU Mem\\(GB)} & \shortstack{Latency\\(ms)} \\
% \midrule
% GSM      & 34.96 & 75.3 & 4832.6 \\
% FlightGPT & 8.29 & 19.6 & 8655.9 \\
% DroneNav  & 7.96 & 16.4 & 4254.7 \\
% \midrule
% \textbf{CityNav-VLA (Ours)}
% & \textbf{4.46}
% & \textbf{8.56}
% & \textbf{679.7} \\
% \bottomrule
% \end{tabular}
% \end{table}

% \input{secv1/5_conclusion}

\section{Conclusion and Limitations}

% CityNav-VLA connects dynamic egocentric context with persistent allocentric memory for language-goal aerial navigation. Metric coordinates provide a stable reference for waypoint prediction, while early goal commitment retains a consistent target during execution. Full-split comparisons show that commitment improves successful arrival, and a controlled RGB intervention confirms the contribution of accumulated visual evidence. Cumulative ablations identify navigation-map features as the largest contributor and show that metric coordinates further improve endpoint localization.

% The current policy uses fixed-horizon continuous control and evaluates arrival at the final position. Extending it to atomic actions and explicit stopping would connect waypoint grounding more closely to task completion. Confidence-based commitment is another promising direction: it could retain the efficiency of a stable target while allowing correction when new observations support a different destination.

CityNav-VLA connects dynamic egocentric context with persistent allocentric memory for language-goal aerial navigation. Metric grounding anchors waypoint prediction, while early goal commitment stabilizes execution. Full-split evaluations and a controlled RGB intervention confirm gains from commitment and accumulated visual evidence. Cumulative ablations identify navigation-map features as the largest contributor, with metric coordinates further improving endpoint localization. However, fixed-horizon control and inflexible commitment limit stopping and target revision. In future work, we will extend CityNav-VLA to support explicit stopping, adaptive goal commitment and lightweight model compression for efficient real-time onboard execution.

% , motivating future work on explicit stopping and adaptive commitment.

\section*{LLM Usage Statement}
ChatGPT was used to assist with manuscript drafting, language editing, text condensation. Moreover, CityBot and Qwen3-VL were used as pretrained multimodal backbones in the navigation method and backbone ablations, as described in the manuscript and supplementary material. The authors reviewed and revised the AI-assisted material and retain full responsibility for the scientific claims, citation accuracy, experimental results, reproducibility, and integrity of the submitted work.

% Flush main-paper figures/tables before references and supplementary material.
% \clearpage
{\small
\bibliographystyle{ieeenat_fullname}
\bibliography{main}
}
% Included after the bibliography; clear queued main-paper floats first.
\clearpage
\maketitlesupplementary
\appendix

% Include this file after the paper's existing \appendix command.
\section{Additional Method Details}
\label{sec:additional_method_details}

\subsection{Input Views and Map Preparation}
\label{sec:additional_map_preparation}

CityBot-4B receives the navigation instruction, local RGB views, and the internally maintained global map view. The local RGB inputs comprise a heading-aligned observation and a north-up crop. The global map view is the accumulated observed RGB layer $A_t$ of $\mathcal{M}_t$.

The map canvas is initialized using CityNav's prior geographic coordinates. The original map is padded to a square and resized to $H\times W$, preserving the correspondence between pixels and physical positions. Canvas position, orientation, and scale remain unchanged throughout an episode. Terrain height, all landmark contours, and instruction-referenced landmark masks are rasterized before navigation and remain fixed geographic layers.

At decision $t$, the current position, height above ground, and heading determine the ground-view footprint $\Omega_t$. Projecting this footprint onto the canvas produces $V_t$, with visible cells marked as 1. The explored mask $E_t$ records the union of all previous visibility masks. The simulator provides a registered overhead RGB image $A^{\mathrm{geo}}\in\mathbb{R}^{H\times W\times3}$; only explored cells are exposed through $A_t=E_t\odot A^{\mathrm{geo}}$. Unseen RGB regions remain masked, while previously revealed regions retain their original map locations.

The full map contains three RGB channels and six single-channel layers: terrain height $H^{\mathrm{geo}}$, all-landmark mask $L^{\mathrm{all}}$, instruction-referenced mask $L^{\mathrm{ref}}$, current visibility $V_t$, explored area $E_t$, and visited positions $P_t$. All nine channels share the same spatial grid. The backbone consumes $A_t$, and the navigation-map encoder consumes the full $\mathcal{M}_t$.

\subsection{Memory Alignment and Fixed Scale Settings}
\label{sec:additional_memory_settings}

Global visual features and navigation-map features are projected onto a common $14\times14$ grid with dimension $D=256$. Corresponding tokens describe the same map region. Learned positional embeddings identify grid cells, and the coordinate branch associates them with physical offsets and distances from the fixed map center.

The coordinate normalization scale $s_c$ is a fixed preprocessing hyperparameter. Its default value is stated in the main text and summarized with the other fixed settings in Table~\ref{tab:method_fixed_settings}. Equation~\ref{eq:metric_coordinate_encoding} expresses the normalization using the symbolic scale. The fusion scalars $\alpha_m$ and $\alpha_c$ are learned jointly with the added navigation modules.

\begin{table}[t]
\centering
\caption{Fixed dimensions and scale hyperparameters.}
\label{tab:method_fixed_settings}
\small
\begin{tabular}{lll}
\hline
Setting & Symbol & Value \\
\hline
Spatial token grid & $N$ & $14\times14=196$ \\
Feature dimension & $D$ & $256$ \\
Coordinate normalization & $s_c$ & $400\,\mathrm{m}$ \\
Vertical normalization & $s_z$ & $100\,\mathrm{m}$ \\
Loss error scale & $\delta$ & $20\,\mathrm{m}$ \\
Vertical loss weight & $\lambda_z$ & $0.25$ \\
\hline
\end{tabular}
\end{table}

\subsection{Decoder and Trainable Components}
\label{sec:additional_trainable_components}

The state MLP and motion-history GRU are projected to the memory dimension and combined with the backbone context to form one navigation-query token. Every decoder block uses $\mathbf{Z}_t$ as keys and values, and contains cross-attention and a feed-forward layer with residual connections. The regression head predicts normalized planar coordinates and ground-relative height; the sigmoid and $\tanh$ output activations apply the bounds specified in Eq.~\ref{eq:waypoint_head}.

CityBot-4B stays frozen. The trainable components are the input projections, state and history encoders, navigation-map and coordinate encoders, positional embeddings, fusion scalars, cross-attention decoder, and regression head. They are optimized jointly with the sample-averaged target-coordinate loss in Eq.~\ref{eq:goal_loss}. EGC is an inference-only target-selection rule and introduces no trainable parameters. Controller settings remain in Appendix~\ref{sec:experimental_details}.

\section{Additional Analyses}
\label{sec:additional_analyses}

\subsection{Qualitative Results}
\label{sec:qualitative}

\begin{figure*}[t]
    \centering
    \includegraphics[width=\textwidth]{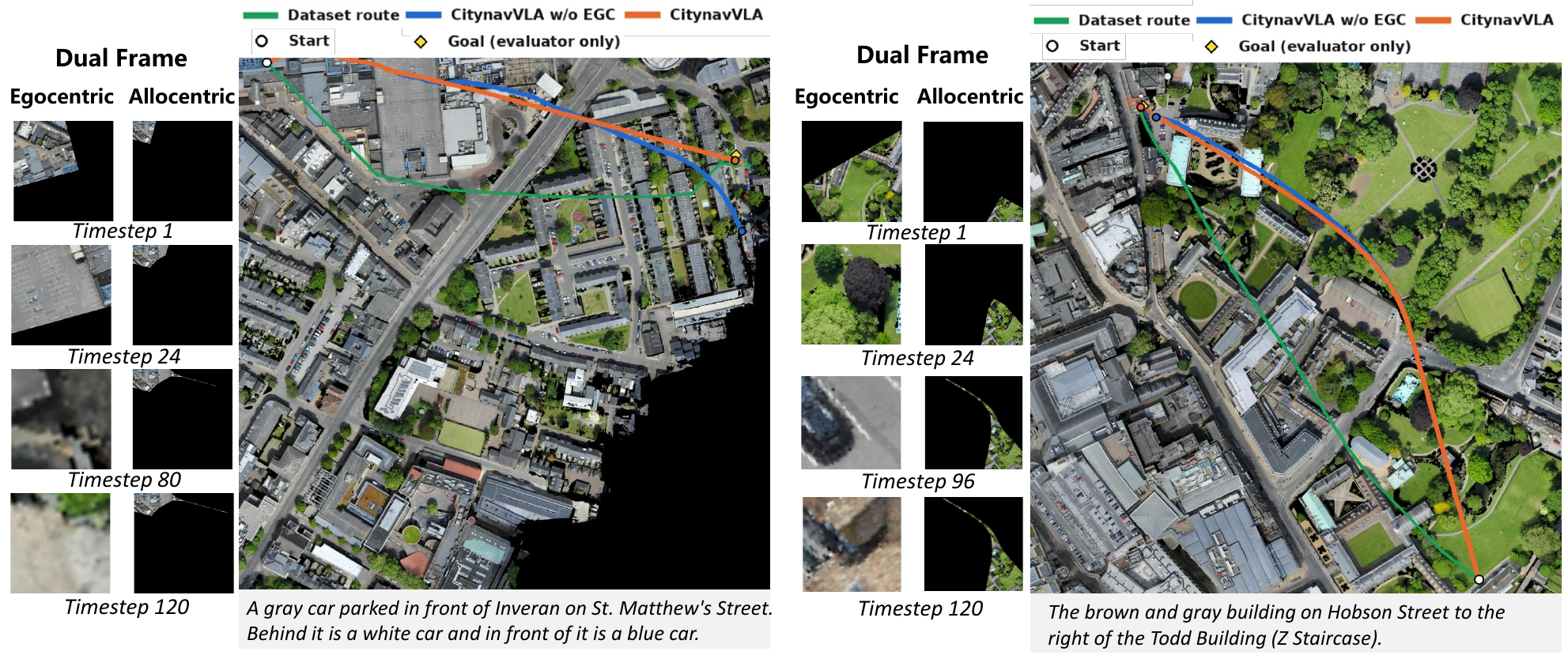}
    \caption{Navigation examples from refined CityNav in map projection. The left panels show observations and map states, while trajectory plots compare execution with and without EGC. Distance annotations are measured in the map plane.}
    \label{fig:qualitative_good}
\end{figure*}

Figure~\ref{fig:qualitative_good} compares navigation with and without EGC in two long-range episodes. The map projections show how commitment preserves an early target estimate while continued replanning can drift. Later observations have smaller ground footprints as the UAV descends. Target markers are added for visualization, and the distance annotations in these projections describe planar geometry.

\begin{figure*}[t]
    \centering
    \includegraphics[width=\textwidth]{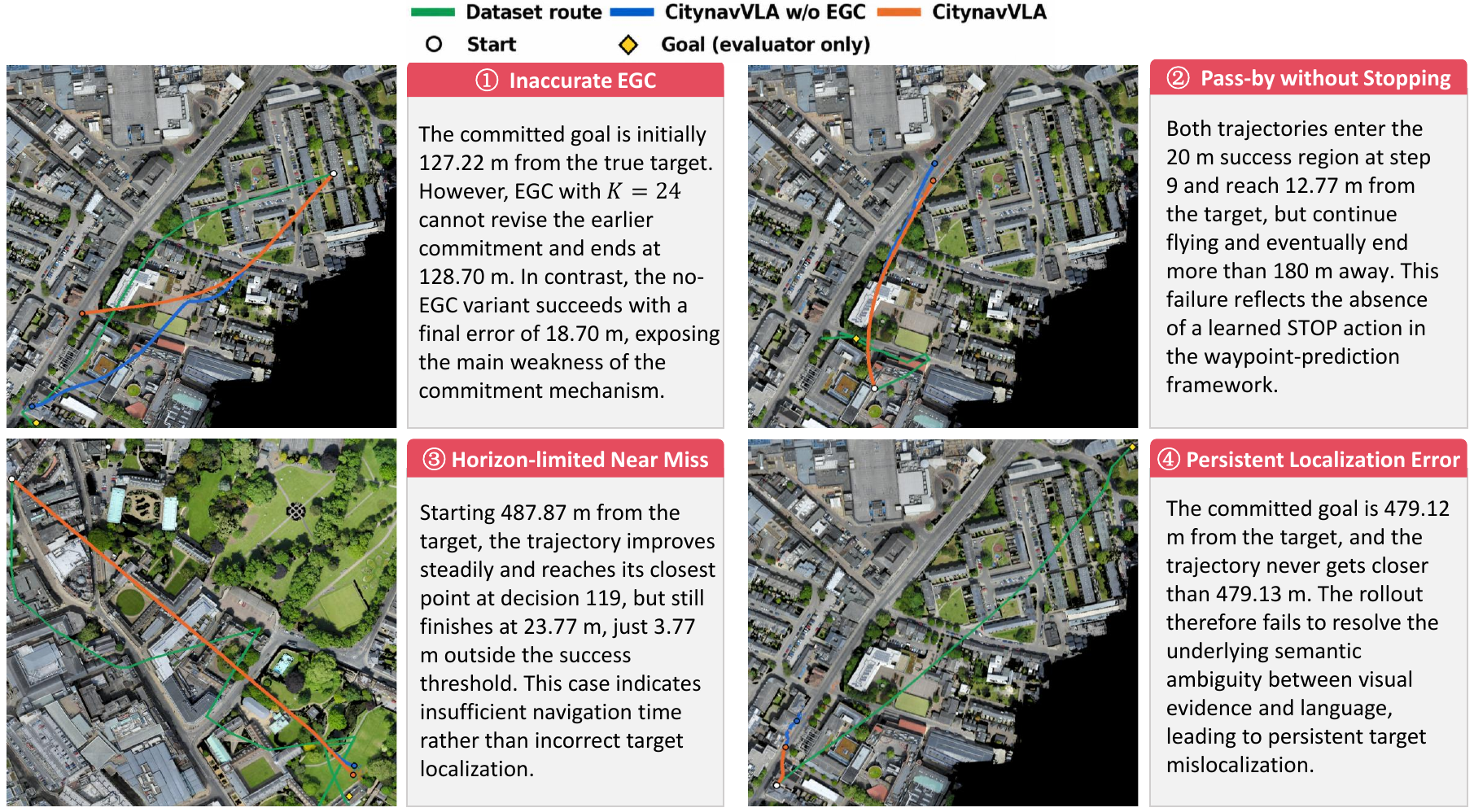}
    \caption{Failure cases from refined CityNav in map projection. Examples show incorrect early commitment, movement away from the target, exhaustion of the decision budget, and localization in the wrong region. Distance annotations are measured in the map plane.}
    \label{fig:qualitative_bad}
\end{figure*}

Figure~\ref{fig:qualitative_bad} shows several characteristic failure modes. Commitment can retain an incorrect early estimate, preventing recovery when later observations reveal a better destination. Without explicit stopping, the UAV can pass near the target and subsequently move away. Other trajectories exhaust the decision budget or remain localized in an incorrect map region, highlighting the importance of reliable goal grounding as well as stable execution.

\subsection{Published Baseline Variants and External Results}
\label{sec:published_difficulty_context}

Seq2Seq+GSM and CMA+GSM in Table~\ref{tab:results} are the GSM-equipped policies reported by the later CityNav evaluation~\cite{lee2025citynav}. GSM augments the inputs with a geographic semantic map. The earlier difficulty results~\cite{lee2024citynav} concern Seq2Seq and CMA without GSM and therefore describe different policy variants. Table~\ref{tab:difficult} presents the difficulty breakdown of our checkpoint. The MGP+LLaVA (HD) row follows the earlier report's map-plane OSR definition.

Table~\ref{tab:geonav_external_subset} reports supplementary GeoNav (DeepSeek) results obtained on a separate server. These completed subsets follow the supplied evaluation convention. Their sample identities and trajectory files are unavailable locally, and the aggregate is presented separately from the full-split results.

\begin{table}[t]
\centering
\caption{Supplementary GeoNav (DeepSeek) results on externally evaluated subsets. Overall is the supplied episode-weighted aggregate. NE is in meters; other metrics are percentages. Sample counts and metric dimensionality were unspecified in the supplied results.}
\label{tab:geonav_external_subset}
\small
\begin{tabular}{@{}lrrrr@{}}
\toprule
Subset & NE$\downarrow$ & SR$\uparrow$ & OSR$\uparrow$ & SPL$\uparrow$ \\
\midrule
Easy & 60.66 & 21.00 & 49.00 & 11.72 \\
Medium & 76.44 & 20.00 & 43.75 & 13.64 \\
Hard & 76.44 & 28.33 & 43.33 & 22.30 \\
Overall & 69.87 & 22.50 & 45.83 & 15.00 \\
\bottomrule
\end{tabular}
\end{table}

\subsection{Effect of Backbone Adaptation}
\label{sec:backbone_adaptation}

We compare frozen raw Qwen3-VL and CityBot-adapted backbones on refined CityNav, keeping the downstream $V+M+P+C$ policy and training schedule fixed. All runs use 10K checkpoints with EGC disabled and cover the complete Validation Unseen and Test Unseen splits. The 4B comparison uses training seeds 42, 43 and 44, while the 8B comparison uses seed 42. Tables~\ref{tab:backbone-4b-full} and \ref{tab:backbone-8b-full} report the two model sizes separately.

% Layout adapted for the supplied CVPR 2027 template; values unchanged.
% Rendered from verified completed records by scripts/paper/render_scis_supplementary_tables.py.
\begin{table*}[t]
\centering
\caption{Backbone adaptation has a small average effect at 4B scale. Scores are means and standard deviations across training seeds 42, 43 and 44 on complete refined CityNav splits. Each run uses a 10K checkpoint without EGC. Scores are planar; NE is in meters and other metrics are percentages.}
\label{tab:backbone-4b-full}
\begingroup
\fontsize{9}{11}\selectfont
\setlength{\tabcolsep}{3pt}
\renewcommand{\arraystretch}{1.13}
\begin{tabular*}{\linewidth}{@{\extracolsep{\fill}}lrrrr@{}}
\toprule
Backbone & NE$\downarrow$ & SR$\uparrow$ & OSR$\uparrow$ & SPL$\uparrow$ \\
\midrule
\multicolumn{5}{@{}l}{\textit{Validation Unseen}} \\
Raw Qwen3-VL-4B & $53.22 \pm 2.65$ & $15.52 \pm 1.71$ & $29.00 \pm 3.72$ & $14.55 \pm 1.77$ \\
CityBot-4B & $52.77 \pm 0.53$ & $15.38 \pm 1.87$ & $28.88 \pm 2.22$ & $14.44 \pm 1.72$ \\
\midrule
\multicolumn{5}{@{}l}{\textit{Test Unseen}} \\
Raw Qwen3-VL-4B & $43.25 \pm 2.82$ & $22.24 \pm 2.55$ & $34.89 \pm 4.34$ & $20.90 \pm 2.49$ \\
CityBot-4B & $43.24 \pm 2.62$ & $22.40 \pm 3.08$ & $34.77 \pm 3.31$ & $20.97 \pm 2.93$ \\
\bottomrule
\end{tabular*}
\endgroup
\end{table*}

\paragraph{4B: effects vary across seeds.}
At 4B scale, adaptation changes average performance only slightly. The SR differences are $-0.15$ and $+0.16$ percentage points on Validation Unseen and Test Unseen, respectively. The direction varies across training seeds, and the differences are small relative to that variation. The three-seed comparison therefore shows similar navigation performance for the two 4B backbones.

\paragraph{8B: endpoint gains with an OSR trade-off.}
At 8B scale, CityBot improves Validation Unseen SR by $2.15$ percentage points ($p=0.0047$) and SPL by $2.05$ points. On Test Unseen, it reduces NE by $1.37\,\mathrm{m}$, while OSR decreases by $2.61$ points. The smaller Test Unseen SR and SPL gains are not statistically significant. Adaptation therefore improves arrival on Validation Unseen and endpoint accuracy on Test Unseen, with reduced trajectory coverage on the latter split.

% Layout adapted for the supplied CVPR 2027 template; values unchanged.
% Rendered from verified completed records by scripts/paper/render_scis_supplementary_tables.py.
\begin{table}[t]
\centering
\caption{Backbone adaptation improves Validation Unseen arrival at 8B scale. Both models use seed 42, 10K checkpoints and the same downstream navigation architecture without EGC. Results cover complete refined CityNav splits and use planar distances. NE is in meters and other metrics are percentages.}
\label{tab:backbone-8b-full}
\begingroup
\fontsize{9}{11}\selectfont
\setlength{\tabcolsep}{2pt}
\renewcommand{\arraystretch}{1.13}
\begin{tabular*}{\linewidth}{@{\extracolsep{\fill}}lrrrr@{}}
\toprule
Backbone & NE$\downarrow$ & SR$\uparrow$ & OSR$\uparrow$ & SPL$\uparrow$ \\
\midrule
\multicolumn{5}{@{}l}{\textit{Validation Unseen}} \\
Raw Qwen3-VL-8B & 53.67 & 14.61 & 30.33 & 13.62 \\
CityBot-8B & 52.80 & 16.76 & 29.55 & 15.67 \\
\midrule
\multicolumn{5}{@{}l}{\textit{Test Unseen}} \\
Raw Qwen3-VL-8B & 43.58 & 22.21 & 37.63 & 20.65 \\
CityBot-8B & 42.21 & 23.04 & 35.01 & 21.74 \\
\bottomrule
\end{tabular*}
\endgroup
\end{table}

Backbone adaptation affects both the spatial tokens and the navigation query. These comparisons measure their joint contribution to the downstream policy, whose response differs between model sizes and evaluation splits.

\subsection{Initial-distance Sensitivity}
\label{sec:distance-analysis}

We examine initial-distance dependence using the refined CityNav main checkpoint without EGC. Episodes are grouped by their initial planar UAV-to-target distance in $25\,\mathrm{m}$ intervals up to $400\,\mathrm{m}$, with longer episodes collected in a final bin. The supplementary data include the number of episodes in each bin.

\begin{figure*}[t]
  \centering
  \includegraphics[width=\textwidth]{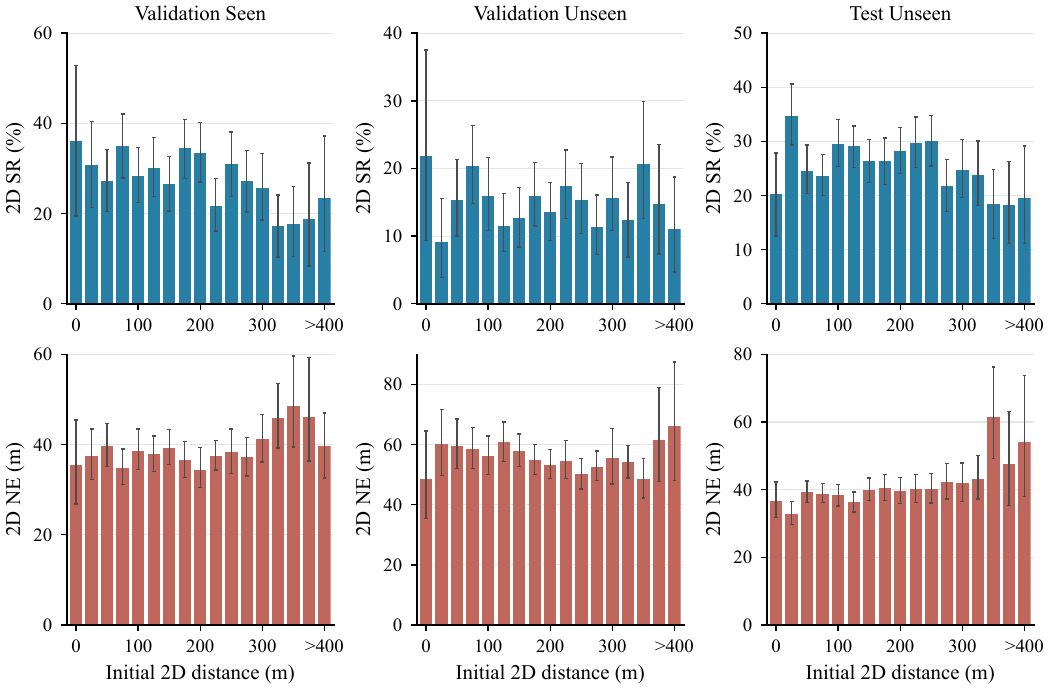}
  \caption{Initial-distance dependence varies across navigation splits. Bars show planar SR and NE on refined CityNav in 25 m intervals; the final hatched bar collects distances beyond 400 m. Error bars show bootstrap confidence intervals.}
  \label{fig:navigation-distance}
\end{figure*}

Longer-distance episodes tend to have lower SR and higher NE on Validation Seen and Test Unseen, whereas Validation Unseen exhibits weaker dependence (Fig.~\ref{fig:navigation-distance}). Initial distance therefore explains only part of the variation in navigation performance. The wider error bars at large distances reflect the smaller number of episodes in those bins.

\subsection{Spatial Reference Representations}
\label{sec:spatial_reference}

We distinguish explicit metric geometry from the reference frame in which it is expressed. The no-coordinate policy retains global visual tokens, navigation-map features and learned positional embeddings. Egocentric encoding describes each map cell relative to the UAV's current position and heading, whereas allocentric $XYR$ uses a fixed map-centered reference. Both metric variants use the same coordinate encoder and absolute target-prediction head.

Table~\ref{tab:spatial-reference} reports the two dataset settings separately. CityNav policies use training seed 44; refined CityNav policies use seed 42, with F2 and F3 from Table~\ref{tab:spatial-memory-components} providing the no-coordinate and allocentric variants. All checkpoints are evaluated after 10K updates under a common 30K learning-rate schedule. The refined comparison additionally tests EGC for each representation.

% Layout adapted for the supplied CVPR 2027 template; values unchanged.
% Rendered from verified completed records by scripts/paper/render_scis_supplementary_tables.py.
\begin{table*}[t]
\centering
\caption{Spatial reference representations on Validation Unseen. CityNav uses seed 44 and 2,826 episodes; refined CityNav uses seed 42 and 2,697 episodes. All policies are evaluated at 10K updates. EGC, when enabled, uses $K=24$. NE is in meters and other metrics are percentages.}
\label{tab:spatial-reference}
\begingroup
\fontsize{9}{11}\selectfont
\setlength{\tabcolsep}{2.5pt}
\renewcommand{\arraystretch}{1.13}
\begin{tabular*}{\linewidth}{@{\extracolsep{\fill}}llrrrr@{}}
\toprule
Representation & EGC & NE$\downarrow$ & SR$\uparrow$ & OSR$\uparrow$ & SPL$\uparrow$ \\
\midrule
\multicolumn{6}{@{}l}{\textit{CityNav, three-dimensional evaluation}} \\
No explicit metric coordinates & Off & 71.24 & 9.55 & 13.48 & 8.97 \\
Egocentric metric displacement & Off & 81.45 & 8.46 & 14.93 & 8.05 \\
Allocentric metric $XYR$ & Off & 69.78 & 9.27 & 14.51 & 8.76 \\
\midrule
\multicolumn{6}{@{}l}{\textit{refined CityNav, planar evaluation}} \\
No explicit metric coordinates & Off & 57.01 & 14.31 & 29.40 & 13.22 \\
Egocentric metric displacement & Off & 61.34 & 11.35 & 27.73 & 10.79 \\
Allocentric metric $XYR$ & Off & 52.42 & 15.76 & 28.03 & 14.68 \\
\midrule
No explicit metric coordinates & On & 56.58 & 14.39 & 27.47 & 13.79 \\
Egocentric metric displacement & On & 52.10 & 14.46 & 29.55 & 13.91 \\
Allocentric metric $XYR$ & On & 51.93 & 16.54 & 27.85 & 15.91 \\
\bottomrule
\end{tabular*}
\endgroup
\end{table*}

On CityNav, allocentric coordinates lower NE from $71.24$ to $69.78\,\mathrm{m}$ relative to the no-coordinate policy. SR and SPL remain similar, with differences of $-0.28$ and $-0.21$ percentage points, respectively, neither statistically significant. The supported gain is more accurate endpoint localization rather than a higher frequency of successful arrival.

On refined CityNav without EGC, allocentric encoding reduces NE by $8.91\,\mathrm{m}$ relative to egocentric displacement and raises SR by $4.41$ percentage points ($p<0.001$). With EGC enabled, allocentric encoding has the highest SR and SPL among the three representations. Its SR gains over no coordinates and egocentric coordinates are $2.15$ and $2.08$ percentage points, respectively, and both are significant.

Commitment has its largest effect on the egocentric policy in this refined comparison, reducing NE by $9.23\,\mathrm{m}$ and raising SR by $3.11$ percentage points. Fixing the execution target reduces the effect of later prediction changes, while the allocentric policy already uses a reference that remains fixed as the UAV moves.

\subsection{Commitment-step Sensitivity}
\label{sec:commitment_step_sensitivity}

The commitment step determines how much evidence the policy gathers before fixing its execution target. We examine $K\in\{8,12,18,24,32,48,64\}$ on all 2,498 Validation Seen episodes using the main CityNav checkpoint. Each setting preserves the first $K$ transitions of the no-EGC rollout and then replays the deterministic controller toward the median of those target predictions. At $K=24$, every replayed position exactly matches its recorded closed-loop counterpart. Scores use three-dimensional distances.

% Layout adapted for the supplied CVPR 2027 template; values unchanged.
% Rendered from verified completed records by scripts/paper/render_scis_supplementary_tables.py.
\begin{table}[t]
\centering
\caption{Early commitment is effective over a range of commitment steps. Scores use three-dimensional distances on all 2,498 Validation Seen episodes. Each setting replays the controller after its unchanged policy prefix; $K=24$ reproduces the recorded closed-loop trajectories exactly. Bold type marks the fixed setting used in the main experiments.}
\label{tab:egc-k-sensitivity}
\begingroup
\fontsize{9}{11}\selectfont
\setlength{\tabcolsep}{2pt}
\renewcommand{\arraystretch}{1.13}
\begin{tabular*}{\linewidth}{@{\extracolsep{\fill}}lrrrr@{}}
\toprule
\shortstack[l]{Commitment\\step $K$} & NE$\downarrow$ & SR$\uparrow$ & OSR$\uparrow$ & SPL$\uparrow$ \\
\midrule
8 & 39.95 & 28.06 & 31.75 & 26.98 \\
12 & 39.96 & 27.86 & 31.71 & 26.76 \\
18 & 39.95 & 28.74 & 32.67 & 27.58 \\
\textbf{24} & 40.08 & 28.38 & 32.31 & 27.20 \\
32 & 40.30 & 28.50 & 32.47 & 27.29 \\
48 & 40.65 & 28.54 & 32.79 & 27.24 \\
64 & 41.12 & 28.70 & 33.11 & 27.31 \\
\midrule
No commitment & 43.10 & 26.14 & 31.14 & 24.72 \\
\bottomrule
\end{tabular*}
\endgroup
\end{table}

Success varies by less than one percentage point across these commitment steps (Table~\ref{tab:egc-k-sensitivity}). The highest observed SR is $28.74\%$ at $K=18$, while the retained setting $K=24$ reaches $28.38\%$. All examined settings improve mean NE and SR over continuous target updates. The results support a range of early commitment steps, with $K=24$ lying within that range. We retain the original setting for both unseen splits.

\section{Experimental Details}
\label{sec:experimental_details}

\paragraph{Dataset and evaluation protocol.}
CityNav~\cite{lee2025citynav} contains 32,637 instruction--trajectory pairs across Training, Validation Seen, Validation Unseen, and Test Unseen splits. Its urban scenes are built from SensatUrban~\cite{hu2022sensaturban}, with landmark locations and contours provided by CityRefer~\cite{miyanishi2023cityrefer}. Refined CityNav, released with HETT~\cite{ding2026hett}, contains 32,326 episodes and revised landmark-description resources. Table~\ref{tab:data_provenance_counts} lists the split sizes. Each dataset's main policy is trained on its corresponding Training Seen split and evaluated on all episodes in each evaluation split.

\begin{table}[htbp]
\centering
\caption{Split sizes of CityNav and refined CityNav.}
\label{tab:data_provenance_counts}
\small
\setlength{\tabcolsep}{3pt}
\begin{tabular}{@{}lrrr@{}}
\toprule
Split & CityNav & refined CityNav & Removed \\
\midrule
Training Seen & 22,002 & 21,878 & 124 \\
Validation Seen & 2,498 & 2,470 & 28 \\
Validation Unseen & 2,826 & 2,697 & 129 \\
Test Unseen & 5,311 & 5,281 & 30 \\
\bottomrule
\end{tabular}
\end{table}

The Easy, Medium, and Hard groups contain $2{,}096$, $1{,}686$, and $1{,}529$ episodes, totaling $5{,}311$. Refined CityNav contains $2{,}081$, $1{,}678$, and $1{,}522$, totaling $5{,}281$. These groups form the two panels of Table~\ref{tab:difficult}; each panel partitions exactly the episodes of its corresponding rows in Table~\ref{tab:results}.

CityNav evaluation uses the original episode files, CityRefer objects and processed descriptions, together with the shared map rasters and landmark geometry. Landmark names are normalized before exact matching. Unmatched names use normalized Levenshtein similarity with a threshold of 0.85 and a margin of 0.05 over the next candidate. All objects sharing a matched name contribute to the landmark mask. Episodes without a resolved reference receive an empty mask: 267 in Validation Seen, 274 in Validation Unseen and 895 in Test Unseen. These episodes remain in the evaluation.

Refined CityNav preserves the instructions, targets, and demonstration trajectories of retained episodes. Its processed landmark annotations contain 18,038 modified records, including capitalization changes, and provide the references used to construct navigation-map features. Sample mappings, resource manifests, and episode-level results are supplied in \path{AuditEvidence/}.

\paragraph{Motion commands and budget.}
The controller first transforms the predicted waypoint displacement into the UAV body frame, giving $(d_f,d_l,d_z)$. With $D$ denoting the three-dimensional distance to that predicted waypoint, it uses
\begin{equation}
\begin{aligned}
\kappa(D)&=12\left[1+2\max\left(0,1-\frac{D}{30}\right)\right],\\
\mathbf{a}_t&=\biggl(
\operatorname{clip}\left(\frac{d_f}{\kappa},-4,4\right),\\
&\qquad\operatorname{clip}\left(\frac{d_l}{\kappa},-4,4\right),\\
&\qquad\operatorname{clip}\left(\frac{d_z}{\kappa},-2,2\right),0\biggr).
\end{aligned}
\label{eq:controller_budget}
\end{equation}
The first three components are displacements in meters and are applied together in one simulator update, while heading remains fixed. Maximum horizontal and XYZ displacement per decision are $4\sqrt{2}$ and 6 m, giving theoretical 120-decision caps of $678.82$ and $720$ m. In refined CityNav's Test Unseen EGC trajectories, 54,231 of 633,720 decisions exceed 5 m horizontally. CityNav applies 5 m forward moves, $30^\circ$ turns, 2 m vertical moves, and STOP as separate actions.

\begin{table*}[t]
\centering
\caption{Action mappings and movement budgets. Travel caps assume 120 atomic decisions for CityNav and 120 continuous updates for CityNav-VLA. Published baseline rows retain their reported horizons, and actual path lengths are measured from executed trajectories.}
\label{tab:action_protocol}
\small
\begin{tabular}{@{}p{.28\linewidth}p{.31\linewidth}p{.33\linewidth}@{}}
\toprule
Property & CityNav discrete actions & Our continuous-control protocol \\
\midrule
One environment decision & One forward, vertical, turning, or STOP action & One simultaneous forward/lateral/vertical displacement \\
Translation & Forward 5 m, or vertical $\pm2$ m & Forward/lateral each in $[-4,4]$ m; vertical in $[-2,2]$ m \\
Heading & Turns of $\pm30^\circ$ consume decisions & Constant; yaw command is zero \\
Largest horizontal / XYZ displacement & 5 / 5 m per decision & $4\sqrt{2}$ / 6 m per decision \\
120-decision horizontal / XYZ cap & 600 / 600 m if all actions are forward & 678.82 / 720 m \\
Termination and SR & STOP within the target sphere & All 120 decisions; final distance at most 20 m in the evaluation coordinate space \\
\bottomrule
\end{tabular}
\end{table*}

On Test Unseen, recorded trajectories with EGC have a mean XYZ path length of $208.16$ m and a maximum of $513.71$ m. On refined CityNav, mean XYZ path lengths are $217.46$ m without EGC and $212.78$ m with EGC, with a maximum EGC length of $512.12$ m.

The shared continuous controller provides a consistent execution setting for the local module and EGC comparisons. Published baseline rows retain the action and stopping settings of their source evaluations.

\paragraph{Metrics and termination.}
For each saved trajectory $\{\mathbf{p}_t\}_{t=0}^{T}$ and dataset target $\mathbf{g}$, $T=120$. The evaluated coordinates are XYZ on CityNav and XY on refined CityNav. All distances, including path length, use the corresponding coordinate space. We use
\begin{equation}
\begin{aligned}
\mathrm{NE}&=\|\mathbf{p}_T-\mathbf{g}\|_2,\\
s&=\mathbf{1}\!\left[\|\mathbf{p}_T-\mathbf{g}\|_2\leq20\,\mathrm{m}\right],\\
o&=\mathbf{1}\!\left[\min_{0\leq t\leq T}\|\mathbf{p}_t-\mathbf{g}\|_2\leq20\,\mathrm{m}\right],\\
d_0&=\|\mathbf{p}_0-\mathbf{g}\|_2,\\
\ell&=\sum_{t=1}^{T}\|\mathbf{p}_t-\mathbf{p}_{t-1}\|_2,\\
\mathrm{SR}&=100\,\mathbb{E}[s],\qquad
\mathrm{OSR}=100\,\mathbb{E}[o],\\
\mathrm{SPL}&=100\,\mathbb{E}\!\left[s\,\frac{d_0}{\max(d_0,\ell,10^{-6})}\right].
\end{aligned}
\label{eq:metrics}
\end{equation}
SR records endpoint arrival, OSR includes the initial position, and SPL uses the initial straight-line distance. The complete XYZ trajectories are retained for both datasets; planar evaluation changes only the reported distance measures, not the controller or trajectories. The dataset target provides training supervision and the reference for trajectory evaluation; policy inputs contain the instruction, observed imagery, navigation map, and UAV state.

\paragraph{Compared methods.}
Table~\ref{tab:results} presents published results under their reported execution settings. HETT$^*$ denotes its refined CityNav evaluation. On refined CityNav, both HETT and CityNav-VLA use planar distances. On CityNav, our scores use XYZ distances, while published baseline values retain their source conventions. Baseline sources and numeric transcriptions are listed in the supplementary material.

\paragraph{Implementation details.}
We use CityBot-4B as the fixed multimodal backbone~\cite{xu2026citycube}. The trainable navigation policy uses a latent dimension of $256$, a $14\times14$ spatial grid, and two cross-attention decoder blocks. Training is performed with AdamW for up to $30{,}000$ steps using a learning rate of $3\times10^{-4}$ and an effective batch size of $8$. We use target-balanced episode sampling and retain the most recent eight states as motion history. The architecture, loss, and execution policy are described in Sec.~\ref{sec:method}.

\paragraph{Checkpoint selection and statistical protocol.}
Both main policies are trained with seed 44. CityNav uses the step-$28{,}000$ checkpoint, selected by Validation Seen closed-loop performance from the fixed 21K and 28K candidates. Its three evaluator runs use seeds 42, 43 and 44 and produce identical episode-level metrics; Table~\ref{tab:results} reports their mean. These runs evaluate one trained policy. Refined CityNav uses the step-$17{,}000$ checkpoint selected by Validation Seen offline waypoint error. Neither unseen split is used for checkpoint selection or hyperparameter tuning.

The cumulative memory study and refined spatial-reference study use seed 42 and 10K checkpoints. The additional CityNav representation study uses seed 44 and the same 10K training duration. Each study retains the learning-rate schedule defined over 30K updates, so the shorter training duration does not accelerate its decay. The CityNav visual-history diagnostic and commitment-step analysis use the main 28K checkpoint. The distance and qualitative analyses retain the refined CityNav main checkpoint.

\paragraph{Visual-history sample selection.}
The 500 Validation Seen episodes are selected before inference by stratifying the full split by scene and initial three-dimensional distance in 100 m intervals. Proportional allocation determines the number drawn from each stratum; sampling within strata uses seed 42 without replacement. The selection does not use model predictions or navigation outcomes. The accumulated-view baseline is taken from the corresponding episodes of the complete no-EGC evaluation.

\paragraph{Paired statistical analysis.}
All intervention comparisons pair the same episodes. We estimate uncertainty in NE and SPL differences using 20,000 bootstrap resamples with seed 42 and retain the central 95\% of the resulting estimates. SR and OSR differences use two-sided exact McNemar tests. Significance is assessed at the 5\% level without adjustment for multiple comparisons. The 4B backbone comparison reports means and standard deviations across its three training seeds and assesses paired differences across seeds separately from episode-level variation.

\end{document}